\documentclass{article}
\usepackage{microtype}
\usepackage{graphicx}
\usepackage{subcaption}
\usepackage{booktabs} 
\usepackage{hyperref}

\PassOptionsToPackage{numbers,compress}{natbib}
\usepackage[main, final]{neurips_2026}

\usepackage{amsmath,amsfonts,bm}

\def\eqref#1{equation~\ref{#1}}

\def\1{\bm{1}}

\def\mA{{\bm{A}}}

\def\mX{{\bm{X}}}

\DeclareMathAlphabet{\mathsfit}{\encodingdefault}{\sfdefault}{m}{sl}
\SetMathAlphabet{\mathsfit}{bold}{\encodingdefault}{\sfdefault}{bx}{n}

\def\gG{{\mathcal{G}}}

\DeclareMathOperator*{\argmin}{arg\,min}

\usepackage{amsmath}
\usepackage{amssymb}
\usepackage{mathtools}
\usepackage{amsthm}
\usepackage{pifont}
\usepackage{enumitem}
\usepackage[most]{tcolorbox}

\usepackage[table,xcdraw,usenames,dvipsnames]{xcolor}

\usepackage[capitalize,noabbrev]{cleveref}
\usepackage{colortbl}
\usepackage{xspace}
\usepackage{siunitx}

\usepackage{listings}
\usepackage{makecell}

\usepackage{url}            
\usepackage{amsfonts}       
\usepackage{nicefrac}       

\usepackage{algorithm}
\usepackage{algpseudocode}

\usepackage{svg}
\usepackage[normalem]{ulem}
\usepackage{dsfont}
\usepackage{multirow}
\usepackage{wrapfig}
\usepackage{array}
\usepackage{float}
\usepackage{marvosym} 

\theoremstyle{plain}
\newtheorem{theorem}{Theorem}[section]
\newtheorem{proposition}[theorem]{Proposition}

\theoremstyle{definition}
\newtheorem{definition}[theorem]{Definition}

\theoremstyle{remark}

\usepackage[disable,textsize=tiny]{todonotes}

\newcommand{\bamo}{BA-2motifs}

\newcommand{\benz}{Benzene}
\newcommand{\mutag}{MUTAG}
\newcommand{\fluo}{Fluoride-Carbonyl}
\newcommand{\alk}{Alkane-Carbonyl}
\newcommand{\ours}{STORE}

\newcommand{\ds}[1]{}

\title{Robust Surrogate Modeling for Explainable Graph Neural Networks}
\author{
  Zhuomin Chen \\
  Florida International University \\
  Miami, FL, USA \\
  \texttt{zchen051@fiu.edu} \\
  \And
  Jingchao Ni \\
  University of Houston \\
  Houston, TX, USA \\
  \texttt{jni7@uh.edu} \\
  \And
  Hojat Allah Salehi \\
  Florida International University \\
  Miami, FL, USA \\
  \texttt{hsalehi@fiu.edu} \\
  \And
  Xu Zheng \\
  Florida International University \\
  Miami, FL, USA \\
  \texttt{xzhen019@fiu.edu} \\
  \And
  Dongsheng Luo\thanks{Corresponding author. This work was primarily conducted while Dongsheng Luo was at Florida International University.} \\
  Singapore Management University \\
  Singapore \\
  \texttt{luodongsheng01@gmail.com} \\
  }

\begin{document}

\maketitle

\begin{abstract}
Graph Neural Networks (GNNs) have achieved remarkable performance in a wide range of graph-related learning tasks. However, explaining their predictions remains a challenging problem, especially due to the mismatch between the graphs used during training and those encountered during explanation. 
Most existing methods optimize soft edge masks on weighted graphs to highlight important substructures, but these graphs differ from the unweighted graphs on which GNNs are trained. This distributional shift leads to unreliable gradients and degraded explanation quality, especially when generating small, sparse subgraphs.
To address this issue, we propose explaining a robust surrogate whose explanations provide a faithful proxy for those of the original model. Specifically, we train this surrogate model through an iterative re-weighting strategy that alternates between identifying explanation subgraphs and adapting the model to the resulting weighted graph distribution, supporting stable explanation refinement while preserving the validity of the soft-mask relaxation.
We evaluate our method on multiple benchmark datasets using different GNN backbones and explanation methods. Experimental results show that our method consistently improves explanation quality and can be flexibly integrated with different architectures.

\end{abstract}

\section{Introduction}
Graph Neural Networks (GNNs) have become a powerful tool for learning from graph-structured data, with applications in molecular property prediction \citep{wu2018moleculenet}, social networks \citep{bu2023improving}, and knowledge graphs \citep{wang2019knowledge}. However, they often behave as black-box models, posing challenges for explainability in high-stakes domains such as healthcare \citep{anklin2021learning} and finance \citep{rai2020explainable}, where transparency is essential.

These concerns have led to the emergence of a number of post-hoc explanation methods to explain predictions of GNNs.
The goal of these methods is to identify a small, label-preserving subgraph that retains the original model prediction $f(G)$ \citep{ying2019gnnexplainer, luo2020parameterized, yuan2020xgnn, wang2022gnninterpreter}. Due to the combinatorial search space, most existing approaches rely on {continuous relaxation}. They introduce a learnable mask $M$, applied to the input graph's adjacency matrix, and optimize $M$ by minimizing the discrepancy between the original prediction $f(G)$ and the masked prediction $f(M \odot G)$. The resulting soft masks are then binarized to yield the final discrete explanation subgraph \citep{yuan2021explainability,vu2020pgm}.

A key assumption underlying this post-hoc explanation framework is that the GNN model $f$, trained exclusively on unweighted graphs, will still produce meaningful predictions and reliable gradients when fed the continuously weighted graphs $M \odot G$ during mask optimization. This assumption is problematic as it introduces a significant {distributional shift}: the model is forced to process graphs with continuous edge weights, which deviate substantially from the binary-weighted graphs in its training distribution. This training scenario reflects the reality of many domains, such as molecular property prediction and social network analysis, where graph connections are often treated as binary (present or absent) due to data constraints or model simplification.

\begin{figure}[t]
    \centering    
    \includegraphics[width=0.8\linewidth]{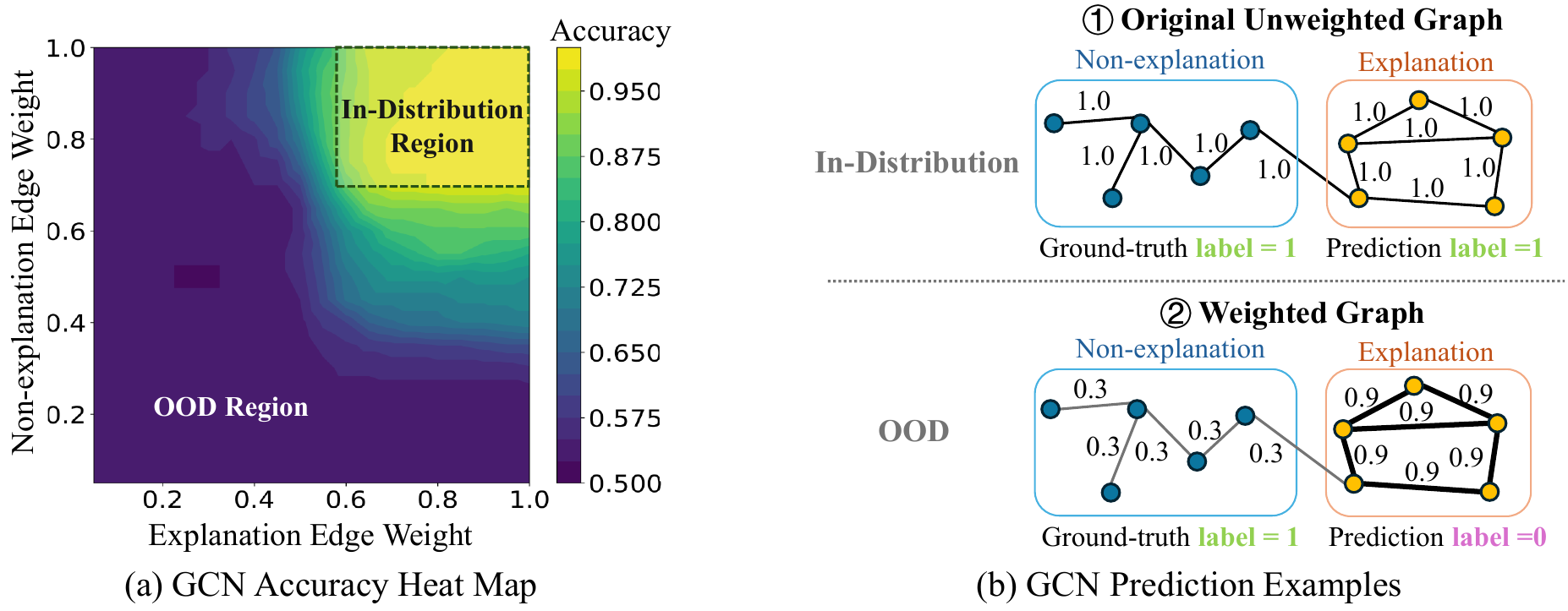}
        \caption{(a) Classification accuracy of a GCN trained on unweighted graphs, evaluated on weighted graphs from the BA-2motif dataset with varying edge weights for motif and BA-graph edges. High accuracy is maintained only in the ``in-distribution'' region (both edge types near weight 1.0), but degrades significantly under distributional shift. (b) The original unweighted graph (example \ding{172}) is correctly classified, but reweighting motif edges to 0.9 and BA-graph edges to 0.3 (example \ding{173}) causes misclassification.}
    \label{fig:introduced-evidence}
\end{figure}

This distributional shift is particularly severe due to the sparsity objective necessary for obtaining faithful explanations. Meaningful explanations typically correspond to small, concise subgraphs, which requires the optimization to push most mask values in $M$ toward zero to satisfy size constraints. This process results in highly sparse, weighted graphs that bear little resemblance to the model's training data. Operating in this Out-Of-Distribution (OOD) regime can cause unstable model behavior and yield unreliable gradient signals, ultimately culminating in suboptimal or unfaithful explanations.
To the best of our knowledge, this explanation-induced distribution shift has not been explicitly identified in prior work on graph explainability.

To empirically validate the impact of distributional shifts, we conduct an experiment on the {\bamo} dataset, which contains graphs with either house or cycle motifs attached to Barabási–Albert (BA) base graphs~\citep{albert2002statistical}.
As shown in Figure~\ref{fig:introduced-evidence}(a), the classification accuracy of a GCN trained on unweighted graphs degrades significantly when evaluated on weighted graphs where edge weights deviate from the training distribution. High accuracy is maintained only in the upper-right ``in-distribution'' region where both motif and base edge weights are near 1.0. Figure~\ref{fig:introduced-evidence}(b) illustrates a concrete failure case: the original unweighted graph is correctly classified (Example \ding{172}), but reweighting edges to reflect a typical explanation configuration (Example \ding{173}) leads to misclassification.
This 
outcome demonstrates that model predictions fail to remain consistent under shifted conditions. If the importance of explanatory edges cannot dominate predictions in this OOD regime, the robustness and trustworthiness of explanations generated by existing methods become highly questionable.

To address this critical limitation, inspired by surrogate explainable AI literature~\citep{ribeiro2016should, hooker2019benchmark}, we propose explaining a \textbf{robust surrogate model} trained to handle continuous edge weights, whose explanations serve as a faithful proxy for those of the original model. We introduce a novel iterative framework 
called graph explanation with stochastic re-weighting ({\ours}) to build and explain this model. Specifically, our method proceeds through alternating phases of {\em subgraph identification} and {\em model adaptation}.
Initially, we extract a relatively large explanation subgraph where soft mask optimization remains reliable. 
We then assign different weights to edges within and outside this subgraph and construct surrogate models by adding these weighted graphs to the training datasets. This step enables surrogate models to maintain predictive consistency under the specific weight distribution that resembles soft-masked graphs, ensuring the validity of subsequent mask optimization.
The process iterates with progressively smaller subgraphs, creating a ``ladder'' in which each step maintains the validity of the soft relaxation assumption. This framework offers several key advantages. First, it directly mitigates the distributional shift problem by ensuring that surrogate models can process weighted graphs at each stage of explanation refinement. Second, the progressive refinement from larger to smaller subgraphs provides a stable optimization trajectory, avoiding the severe distributional shifts that occur when directly searching for minimal explanations. Third, it naturally produces explanations at multiple levels of granularity, offering practitioners flexibility in balancing explanation size with fidelity. We summarize our contributions as follows: 

\begin{itemize}[leftmargin=*]
\setlength{\itemsep}{-0.5pt}
    \item We identify and formally characterize a critical flaw in post-hoc GNN explainability: the distributional shift between unweighted training graphs and the weighted graphs used during mask optimization, which undermines explanation reliability.
    \item We propose a robust surrogate framework designed to resolve this OOD problem. By constructing
    a surrogate model capable of processing continuous edge weights, our framework ensures reliable gradient estimation and preserves the validity of soft mask optimization.
    \item We instantiate this framework through a novel iterative stochastic re-weighting strategy. This approach progressively aligns the surrogate model's training distribution with the explanation setting by assigning importance-aware weights to explanatory and non-explanatory edges.
    \item Extensive experiments on five benchmark datasets demonstrate that our method consistently improves explanation quality with different GNNs and explanation methods, while incurring only modest computational overhead.
\end{itemize}
\section{Preliminaries and the OOD Problem}
We represent a graph as $G=( \mathcal{V}, \mathcal{E}, \mX, \mA)$, where $\mathcal{V} = \{v_1, v_2, \dots, v_n\}$ is the set of nodes with $n = |\mathcal{V}|$, $\mathcal{E} \subseteq \mathcal{V} \times \mathcal{V}$ is the set of edges, $\mX \in \mathbb{R}^{n \times d}$ is the node feature matrix where the $i$-th row $\mX_i$ represents the $d$-dimensional feature of node $v_i$, and $\mA \in \{0,1\}^{n \times n}$ is the binary adjacency matrix such that $A_{i,j} = 1$ if $(v_i, v_j) \in \mathcal{E}$. Each graph is associated with a label $ y \in \mathcal{Y} $.

Let $f:\gG\mapsto \{1,2,\cdots,|\mathcal{Y}|\}$ denote a to-be-explained GNN model that maps an input graph $G\in \gG$ to its predicted class. This work focuses on \textit{post-hoc, instance-level} explanation methods, which treat the GNN as a black box and aim to provide task-agnostic explanations for its predictions \citep{ying2019gnnexplainer, luo2020parameterized, yuan2022explainability, huang2024factorized}. 
Given a GNN model $f$ and the input graph $G$, the purpose of the explainer is to find a subgraph $G_{exp} \subseteq G$ containing relevant information for the model's prediction while excluding unnecessary structure. We evaluate an explanation based on two properties. Sufficiency measures how closely the predictive distribution conditioned on $G_{exp}$ agrees with that conditioned on the full graph $G$, expressed as $d_{TV}(P_{\widehat{Y}|G}, P_{\widehat{Y}|G_{exp}\subseteq G}) \approx 0$. Minimality encourages a compact explanation by constraining the number of selected edges. Following the sufficiency–minimality formulation introduced in \citep{chen2026addressing}, our objective is to derive an explainer $\Psi(\cdot)$, with $G_{exp} = \Psi(G)$, to minimize the discrepancy of the two predictive distributions based on the expectation of an edge budget $s$:
\begin{align}
\label{eq:exp_opt}
\begin{aligned}
  \Psi^*
  &= \argmin_{\Psi}
     \; \mathbb{E}_G \!\left[
       d_{TV}\!\left(f(G), P_{\widehat{Y}\mid G_{exp}\subseteq G}\right)
     \right], 
  &\text{s.t.} \quad
     \mathbb{E}_G[|\mathcal{E}_{exp}|] \leq s.
\end{aligned}
\end{align}
Here, $f(G)$ represents the predictive distribution $P_{\widehat{Y}|G}$. The remaining conditional distribution, $P_{\widehat{Y}|G_{exp}\subseteq G}$, is generally difficult to evaluate. This makes direct optimization impractical and motivates the approximations used in existing explanation methods.

To practically solve the optimization problem in \eqref{eq:exp_opt}, existing methods learn a parameterized explainer function, denoted as $\Psi_\psi$, where $\psi$ represents the learnable parameters. For a given graph $G$, this function generates an intermediate, weighted subgraph, $\Psi_\psi(G)$, where edge weights signify their importance to the model's prediction. The final discrete explanation, $G_{exp}$, is then typically derived by binarizing this weighted subgraph.

The standard approach in the literature is to optimize the parameters $\psi$ by computing the sufficiency loss on this intermediate weighted subgraph. The formal objective from ~\eqref{eq:exp_opt} is thus approximated by an objective over $\psi$. The sufficiency term is replaced by a proxy like the Cross-Entropy (CE) loss, and the minimality constraint is handled by a regularization term that encourages sparsity in the generated subgraph, such as the norm of its weighted adjacency matrix, $||\Psi_\psi(G)||$. This leads to the following standard optimization objective:
\begin{equation}
\label{eq:previous_obj}
\begin{aligned}
\psi^* &= 
&\argmin_{\psi} \Bigg (
    \sum_{(G, y) \in \mathcal{G}} \mathrm{CE}\!\big(f(G), f(\Psi_\psi(G))\big)
    + \lambda \|\Psi_\psi(G)\|
\Bigg).
\end{aligned}
\end{equation}
However, we claim this standard formulation contains a critical flaw. The GNN model $f$ is typically trained on graphs with unweighted edges (i.e., binary adjacency matrices). The intermediate subgraph, $\Psi_\psi(G)$, is inherently weighted. Consequently, evaluating the original model $f$ on this subgraph constitutes an OOD problem. The gradients produced in this regime, which are used to update the explainer's parameters $\psi$, may be unstable and unreliable.

\begin{figure*}[t]
    \centering    
    \includegraphics[width=1.0\textwidth]{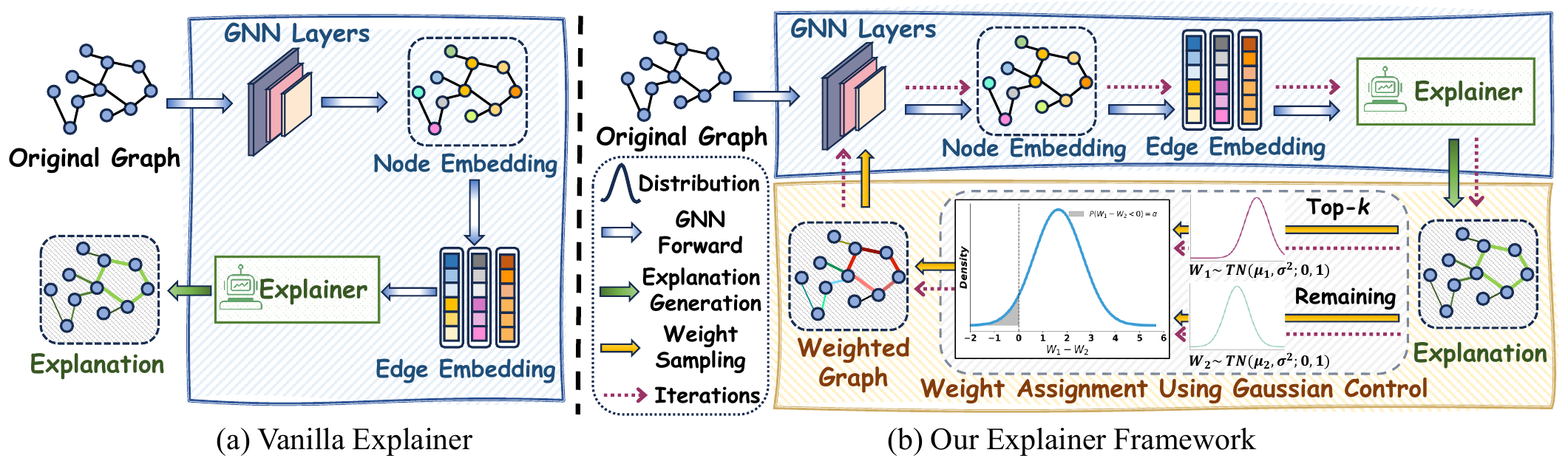}
    \caption{Overview of the vanilla explainer and \ours. (a) Vanilla explainer. (b) {\ours} first inputs the original graph into a GNN to obtain node and edge embeddings. Then we use an explainer to generate a weighted explanation subgraph. Based on this explanation, we apply a Gaussian-based sampling strategy to assign edge weights and construct a weighted graph. This graph is used to retrain a surrogate GNN to improve explanation accuracy.} 
    \label{fig:model}
\end{figure*}

\section{Robust Surrogate Framework}
In this section, we present our graph explanation with stochastic re-weighting ({\ours}) framework. An overview is shown in Figure~\ref{fig:model}(b). 
Our approach addresses the distributional shift problem by iteratively building a \textbf{robust surrogate model} through a process that alternates between {\em subgraph identification} and {\em model adaptation}.
Given a pre-trained GNN model, we first extract a relatively large explanation subgraph where soft mask optimization remains reliable. We then construct weighted training graphs by assigning different weights to edges within and outside the identified subgraph, and construct a surrogate GNN on these weighted graphs to ensure predictive consistency under the corresponding weight distribution. This process iterates for $L$ rounds, progressively refining explanations from larger to smaller subgraphs. 
At each iteration, the explainer identifies important substructures, while the surrogate GNN maintains consistent predictions on weighted graphs, thereby preserving the validity of continuous relaxation throughout the explanation refinement process.


\subsection{Explanation-Guided Stochastic Re-Weighting}

Our iterative process begins by generating a foundational set of explanations to guide model adaptation. To accomplish this, we initialize $f^{(0)}$ with $f$, where $f$ is the pre-trained GNN model to be explained, and train an initial explainer, $\Psi_\psi^{(0)}$ on $f^{(0)}$ using the standard objective from~\eqref{eq:previous_obj}. This process yields a continuous score for each edge, representing its learned importance. From these scores, we derive an initial explanation subgraph by selecting the top-$k$ edges. Following a common strategy, we use a relatively large `$k$' in this first step to create an inclusive set of potentially important edges.

This initial subgraph provides the necessary input for our re-weighting strategy. However, since the explanation is generated under the previously discussed OOD condition, the core of our method is to address this distributional shift directly. We achieve this by training a surrogate GNN model on weighted graphs constructed through a process we call \textbf{stochastic re-weighting}. This adaptation step supervises the surrogate model to maintain consistent predictions when processing graphs with varying edge weights, which is a critical requirement for generating reliable explanations.

As illustrated in Figure~\ref{fig:model}(b), given explanation subgraphs from the initial explainer, we construct weighted training graphs by stochastically assigning different weights to explanatory and non-explanatory edges indicated by the explanation subgraphs. Specifically, we use two well-separated truncated Gaussian distributions: explanatory edges receive weights sampled from a higher-mean distribution, while non-explanatory edges are sampled from a lower-mean distribution. This stochastic re-weighting strategy ensures that the surrogate model learns to respect the relative importance of explanatory edges, and the controlled randomness helps it generalize to the continuous weight variations encountered during soft mask optimization.

\noindent \textbf{Gaussian Parameter Selection.} Our stochastic re-weighting strategy samples edge weights to construct new weighted training graphs. Specifically, we use two truncated Gaussian distributions: explanatory edges receive weights sampled from $W_1 \sim \mathcal{TN}(\mu_1, \sigma^2; 0, 1)$, while non-explanatory edges receive weights from $W_2 \sim \mathcal{TN}(\mu_2, \sigma^2; 0, 1)$. We use the $[0, 1]$ interval because this range directly mimics the continuous, weighted graphs generated by the soft mask optimization in standard GNN explainers, which is essential for bridging the distributional shift.

To parameterize these distributions, we first fix two hyperparameters: the desired mean separation $\Delta \mu = \mu_1 - \mu_2 > 0$ and the overlap probability $\alpha = \mathbb{P}(\tilde{W}_1 < \tilde{W}_2)$, where $\tilde{W}_1 \sim \mathcal{N}(\mu_1, \sigma^2)$ and $\tilde{W}_2 \sim \mathcal{N}(\mu_2, \sigma^2)$ are the untruncated counterparts of $W_1$ and $W_2$. We set $\alpha=0.001$ to ensure a clear separation between the two edge types. The overlap probability is given by:
$\mathbb{P}(\tilde{W}_1 < \tilde{W}_2) = \mathbb{P}(\tilde{W}_1 - \tilde{W}_2 < 0) = \Phi\left( \frac{-\Delta \mu}{\sqrt{2\sigma^2}} \right)$,
where $\Phi(\cdot)$ is the standard normal cumulative distribution function: $\Phi(w) = \frac{1}{\sqrt{2\pi}} \int_{-\infty}^w e^{-t^2 / 2} \, dt$.
By fixing $\Delta \mu$ and $\alpha$, the variance $\sigma^2$ is computed as:
$\sigma^2 = \frac{(\Delta \mu)^2}{2\left[\Phi^{-1}(\alpha)\right]^2}$.
Once $\sigma$ is determined, we select the means. We sample $\mu_2$ from $\text{Uniform}(2\sigma, 1 - \Delta \mu - 2\sigma)$, which in turn determines $\mu_1 = \mu_2 + \Delta\mu$. This sampling strategy ensures that both $\mu_1$ and $\mu_2$ are positioned at least $2\sigma$ from the $0$ and $1$ boundaries. Consequently, the probability of the untruncated Gaussians $\tilde{W}_1$ and $\tilde{W}_2$ falling outside $[0, 1]$ is negligible. This alignment ensures that the truncated distributions $\mathcal{TN}$ used in practice retain means and variances almost identical to their untruncated $\mathcal{N}$ counterparts, preserving the desired separation properties.

\noindent \textbf{Constructing Weighted Training Graphs.} 
For each graph $G$ in the training set, we use the current explainer $\Psi$ to identify the set of explanatory edges $\mathcal{E}^{\text{exp}}$. We then construct a weighted graph $G'$ by assigning edge weights sampled from $\mathcal{TN}(\mu_1, \sigma^2; 0, 1)$ to explanatory edges and from $\mathcal{TN}(\mu_2, \sigma^2; 0, 1)$ to non-explanatory edges. This stochastic re-weighting creates a training distribution that bridges the gap between unweighted graphs and the weighted graphs encountered during mask optimization. The full process is detailed in Appendix \ref{app:algorithm} Algorithm \ref{alg:aug_graphs}.

The surrogate GNN is then trained on both the original graphs $\mathcal{G}$ and the weighted graphs $\mathcal{G}'$:
\begin{equation}
\argmin_{f^{(l)}} \left( \sum_{(G, y) \in \mathcal{G} \cup \mathcal{G}'} -y \log(f^{(l)}(G)) \right),
\end{equation}
This joint training ensures that the surrogate model maintains accurate predictions on both unweighted graphs (preserving original performance) and weighted graphs (enabling reliable mask optimization).

\subsection{Iterative Refinement Process}

Our framework alternates between training the surrogate GNN on weighted graphs and updating the explainer, progressively refining explanations over $L$ iterations. After updating the surrogate at each iteration, we update the explainer using the corresponding surrogate model
$f^{(l)}$. This iterative process, summarized in Appendix~\ref{app:algorithm} Algorithm~\ref{alg:process}, progressively refines from larger to smaller subgraphs, which alleviates distributional shifts that occur when 
searching for minimal explanations. 

\noindent \textbf{Remark.} Despite involving iterative surrogate refinement,
the computational overhead remains modest. In practice, we find that the framework requires only a small number of iterations ($\leq 3$) , and significant improvements are often observable after just a single adaptation step. As empirically demonstrated in Section~\ref{sec:efficiency} Figure~\ref{fig:time_cost}, the cost per iteration is low, confirming that the iterative refinement process is computationally efficient.
\section{Theoretical Analysis}
Our framework is designed to address the critical issue of distributional shift, where a GNN trained on unweighted graphs yields unreliable explanations when evaluated on the weighted graphs inherent to the explanation process. The iterative refinement process, which produces a final robust model, $f^{(L)}$, is empirically effective. Here, we provide the theoretical foundation that justifies when explanations from our final robust model can be considered faithful to the original, non-robust model, $f^{(0)}$.

Following \citet{zheng2024towards}, a task is $(s,\kappa)$-explainable if there exists an explanation function $\Psi$ such that the label is nearly conditionally independent of the full input given $\Psi({G})$, with expected explanation size at most $s$. A classifier is $(s,\zeta)$-explainable under the same condition with predictions replacing labels. We also adopt the stability assumption from \citet{zheng2024towards}, which requires that if $\Psi({g}) \subseteq {g}'$, then $\Psi({g}') = \Psi({g})$. This assumption ensures that explanations remain consistent when the input graph is extended to a supergraph. Under this uniqueness condition, we can formally connect the explanations of the original model $f^{(0)}$ and the robust model $f^{(L)}$. 
The following proposition establishes that,
when both models achieve errors close to the Bayes optimal, they share the same explanation function almost surely.


\begin{proposition}[Shared explanation for $f^{(0)}$ and $f^{(L)}$]
\label{prop:storeshared}
Consider a binary task with Bayes error $\epsilon^\star$. If the task is uniquely $(s,\kappa)$-explainable by $\Psi^\star$ under the stability assumption, and if $f^{(0)}$ and $f^{(L)}$ achieve errors at most $\epsilon^\star+\delta_0$ and $\epsilon^\star+\delta_L$, then both classifiers are $(s,\eta_i)$-explainable with respect to $\Psi^\star$ (with $\eta_i \to 0$ as $\delta_i,\epsilon^\star,\kappa \to 0$) and share the same explanation almost surely.
\end{proposition}

This result shows that,
under the stability and uniqueness assumptions, the robust model $f^{(L)}$ and the original model $f^{(0)}$ share the same underlying explanation almost surely. In other words, explanations from the robust model can be viewed as faithful surrogates for those of the original model, with the approximation error shrinking as both models approach Bayes optimality. The assumptions themselves are natural in many settings: stability reflects the idea that once a subgraph is sufficient to explain a prediction, adding irrelevant structure should not change the explanation; uniqueness means that for each instance, there is a well-defined explanatory subgraph. These principles echo the intuition behind widely used explanation methods such as LIME~\citep{ribeiro2016should}, which rely on the existence of a consistent local explanation. Together, the assumptions highlight why our theoretical result is relevant in practice.
This is further supported by the empirical analysis in Appendix~\ref{app:validity}, where we show that the robust surrogate preserves both the predictive behavior and explanatory structures of the original GNN.

\section{Related Work}

\noindent\textbf{GNN Explanation Methods.}
Existing research on GNN explainability primarily focuses on instance-level explanations, where the goal is to identify influential subgraphs for specific predictions. The dominant paradigm relies on perturbation, where methods optimize soft or binary masks to isolate subgraphs that preserve the original prediction. This category includes approaches that learn soft edge masks via optimization \citep{ying2019gnnexplainer, luo2020parameterized, luo2024towards, wang2021towards}, employ search-based policies \citep{yuan2021explainability, li2023dag}, or leverage game-theoretic values \citep{zhang2022gstarx}. 
Recent studies have further advanced GNN explainability through factorized explanation models \citep{huang2024factorized}, graph output attribution \citep{lu2024goat}, efficient edge-induced subgraph search \citep{lu2024eigsearch}, and structure-aware interaction indices \citep{pmlr-v235-bui24b}. Other work has extended graph interpretability to specialized settings, including 3D molecular graphs \citep{pmlr-v267-qu25a} and multi-granular graph representations \citep{yang2025from}.
Despite improving fidelity, these methods inherently force the GNN to process modified graph structures that deviate from the discrete graphs seen during training, introducing artificial evidence.

\noindent\textbf{Distribution Shift in GNN Explanations.}
To mitigate distribution shifts in GNN explanations, existing studies have explored various strategies for generating in-distribution explanations. MixupExplainer \citep{zhang2023mixupexplainer} uses graph augmentation to reduce distribution shifts, while ProxyExplainer \citep{chen2024generating} constructs in-distribution proxy graphs to facilitate explanation generation. RegExplainer \citep{zhang2024regexplainer} further adopts a mixup-based strategy to address distribution shifts in graph regression explanations. Other approaches, such as D4Explainer \citep{chen2023d4explainer} and Gen-GraphEx \citep{saha2025gen}, leverage generative models to produce in-distribution graph explanations, primarily for counterfactual or model-level interpretability. However, these approaches do not address the specific mismatch identified in our work, where the target GNN is trained on unweighted graphs but explained using continuous weighted masks.
Unlike these methods, our {\ours} framework explicitly bridges this gap by training a robust surrogate model capable of handling the continuous edge weights inherent to the explanation process.

\section{Experiments}
\subsection{Experimental Setup}
\noindent \textbf{Datasets.} We evaluate our method on five benchmark datasets with ground-truth explanations. These include four real-world datasets: {\mutag}~\citep{luo2020parameterized}, {\benz}, {\alk}, and {\fluo}~\citep{agarwal2023evaluating}, as well as one synthetic dataset, {\bamo}~\citep{luo2020parameterized}. Detailed dataset statistics and descriptions are provided in Appendix \ref{app:data}. 


\noindent \textbf{Baselines.}
We compare our method with five representative explainers: GNNExplainer~\citep{ying2019gnnexplainer}, PGExplainer~\citep{luo2020parameterized}, ReFine~\citep{wang2021towards}, MixupExplainer~\citep{zhang2023mixupexplainer}, and ProxyExplainer~\citep{chen2024generating}. 
We adopt a three-layer Graph Convolutional Network (GCN)~\citep{kipf2017semisupervised} as the backbone, using the experimental settings in \citep{ying2019gnnexplainer, luo2020parameterized}.
Experiments on another representative GNN, Graph Isomorphism Network (GIN)~\citep{xu2018powerful}, are provided in Appendix \ref{app:baseline}. 

\begin{table*}[t]
    \centering
    \small
    \caption{Performance comparison of different explanation methods using GCN as the backbone}
    \begin{tabular}{m{2.5cm}|>{\centering\arraybackslash}m{1.8cm}>{\centering\arraybackslash}m{1.8cm}>{\centering\arraybackslash}m{1.8cm}>{\centering\arraybackslash}m{1.8cm}>{\centering\arraybackslash}m{1.8cm}}
    \hline
    Explainer Method & {\bamo} & {\mutag} & {\benz} & Alkane-Car.  & Fluoride-Car.  \\ \hline
    GNNExplainer   & 0.515$_{\pm 0.107}$ & 0.707$_{\pm 0.086}$ & \textbf{0.604$_{\pm 0.029}$} & 0.650$_{\pm 0.042}$ & 0.640$_{\pm 0.018}$  \\ 
                           {\scriptsize \hfill + \ours}    & \textbf{0.528$_{\pm 0.108}$} & \textbf{0.720$_{\pm 0.057}$} & 0.595$_{\pm 0.026}$ & \textbf{0.686$_{\pm 0.026}$} & \textbf{0.647$_{\pm 0.019}$} \\ 
    \hline
    PGExplainer  & 0.672$_{\pm 0.326}$ & 0.533$_{\pm 0.272}$ & 0.541$_{\pm 0.204}$ & 0.778$_{\pm 0.237}$ & 0.784$_{\pm 0.031}$ \\ 
                           {\scriptsize \hfill + \ours}  & \textbf{0.758$_{\pm 0.232}$}  & \textbf{0.774$_{\pm 0.189}$} & \textbf{0.662$_{\pm 0.152}$} & \textbf{0.801$_{\pm 0.227}$} & \textbf{0.802$_{\pm 0.020}$}  \\ 
    \hline
    Refine & 0.522$_{\pm 0.078}$ & 0.502$_{\pm 0.193}$ & 0.723$_{\pm 0.166}$ & 0.623$_{\pm 0.268}$ & 0.633$_{\pm 0.051}$  \\ 
                           {\scriptsize \hfill + \ours}  & \textbf{0.546$_{\pm 0.095}$}  & \textbf{0.529$_{\pm 0.197}$} & \textbf{0.747$_{\pm 0.178}$} & \textbf{0.752$_{\pm 0.153}$} & \textbf{0.648$_{\pm 0.043}$}  \\ 
    \hline
    MixupExplainer & 0.847$_{\pm 0.121}$ & \textbf{0.799}$_{\pm 0.154}$ & 0.555$_{\pm 0.270}$ & 0.444$_{\pm 0.312}$ & 0.664$_{\pm 0.090}$  \\ 
                           {\scriptsize \hfill + \ours}  & \textbf{0.871$_{\pm 0.110}$}  & 0.783$_{\pm 0.160}$ & \textbf{0.570$_{\pm 0.252}$} & \textbf{0.493$_{\pm 0.299}$} & \textbf{0.667$_{\pm 0.130}$}  \\ 
    \hline
    ProxyExplainer & 0.729$_{\pm 0.226}$ & 0.614$_{\pm 0.232}$ & 0.584$_{\pm 0.214}$ & 0.678$_{\pm 0.198}$ & 0.718$_{\pm 0.036}$  \\ 
                           {\scriptsize \hfill + \ours}  & \textbf{0.812$_{\pm 0.137}$}  & \textbf{0.615$_{\pm 0.233}$} & \textbf{0.615$_{\pm 0.190}$} & \textbf{0.766$_{\pm 0.128}$} & \textbf{0.742$_{\pm 0.029}$}  \\ 
    \hline
    \end{tabular}
    \label{tab:exp_gcn}
\end{table*}

\begin{table*}[t]
\centering
\small
\caption{Fidelity evaluation}
\label{tab:fidelity}
\begin{tabular}{p{1.0cm}p{1.2cm}>{\centering\arraybackslash}m{1.7cm}>{\centering\arraybackslash}m{1.7cm}>{\centering\arraybackslash}m{1.7cm}>{\centering\arraybackslash}m{1.7cm}>{\centering\arraybackslash}m{1.7cm}}
\toprule
 & & {\bamo} & {\mutag} & Fluoride-Car. & {D\&D} & {Graph-SST2} \\
\midrule
\multirow{2}{*}{\small $Fid_{\alpha_1,{+}} \uparrow$} 
    & PGExplainer & 0.038$_{\pm 0.029}$ & 0.042$_{\pm 0.008}$ & 0.055$_{\pm 0.012}$ & 0.010$_{\pm 0.011}$ & 0.008$_{\pm 0.002}$\\
    & {\scriptsize \hfill + \ours} & \textbf{0.074$_{\pm 0.023}$} 
                               & \textbf{0.060$_{\pm 0.002}$} 
                               & \textbf{0.064$_{\pm0.010}$} & \textbf{0.022$_{\pm 0.008}$} & \textbf{0.011$_{\pm 0.001}$} \\
\hline
\multirow{2}{*}{\small $Fid_{\alpha_2,{-}} \downarrow$} 
    & PGExplainer & 0.037$_{\pm 0.023}$ & 0.063$_{\pm 0.044}$ & 0.018$_{\pm 0.007}$ & 0.003$_{\pm 0.005}$ & 0.005$_{\pm 0.002}$ \\
    & {\scriptsize \hfill + \ours} & \textbf{0.035$_{\pm0.022}$} 
                               & \textbf{0.041$_{\pm 0.022}$} 
                               & \textbf{0.016$_{\pm 0.008}$} & \textbf{0.001$_{\pm 0.004}$} & \textbf{0.003$_{\pm 0.002}$} \\
\bottomrule
\end{tabular}
\end{table*}

\noindent \textbf{Settings.}
Unlike previous works~\citep{ying2019gnnexplainer,luo2020parameterized}, which train a single GNN with a fixed seed and evaluate the explainer using 10 random seeds, we observe that the robustness of the GNN itself can vary substantially across different initializations and may exhibit different degrees of distributional shift. To obtain a more reliable and unbiased evaluation, we train both the GNN and the explainer under 10 random seeds. For each seed, a GNN is trained from scratch and an explainer is applied to that model, and we report the final results as the average across the 10 runs\footnote{Our code is available at \url{https://github.com/realMoana/STORE}}.
Additional implementation details are provided in Appendix \ref{app:implementation}.

\subsection{Quantitative Evaluation}
\label{sec:exp:quantitative}
We report the mean AUC-ROC over 10 random runs to quantitatively evaluate the quality of explanations. Table~\ref{tab:exp_gcn} shows the results for five explanation methods, each evaluated in both their vanilla form and when integrated with our iterative framework, using GCN as the backbone. From the results, we have several key observations: 
First, although different explanation methods exhibit varying levels of performance, all of them consistently benefit from our iterative framework, \ours, which improves AUC-ROC scores over their vanilla counterparts across most datasets. 

For example,
on {\bamo}, we observe a 12.8\% relative gain for PGExplainer and a 11.4\% improvement for ProxyExplainer, confirming the generalizability of our method across both synthetic and real-world datasets. 
Finally, \ours\ enhances stability, particularly on OOD-prone datasets such as {\bamo}. While vanilla explainers (e.g., MixupExplainer) experience instability in some datasets, our method mitigates this by preserving semantically important substructures during iterative training. This targeted perturbation reduces the risk of degrading important graph structures, resulting in more reliable explanations.
Moreover, Appendix \ref{app:gin} reports results under the GIN architecture, showing that \ours\ is compatible with different GNN architectures and explainers.

To further evaluate the faithfulness of explanations, we adopt fidelity-based metrics. While standard fidelity metrics~\citep{yuan2021explainability} are widely used, they suffer from OOD problem~\citep{amara2023ginx}, which compromise their reliability. To address this, we use the robust fidelity scores ($Fid_{\alpha_1,{+}}$, $Fid_{\alpha_2,{-}}$), following the formulation in~\citep{zheng2024towards}, with the default parameters $\alpha_1=0.1, \alpha_2=0.9$. As shown in Table~\ref{tab:fidelity}, {\ours} 
outperforms all baselines under both metrics, including the large-scale D\&D~\citep{dobson2003distinguishing} and Graph-SST2 datasets~\citep{yuan2022explainability}.
We also provide a visual comparison in Appendix \ref{app:casestudy}, including the ground-truth explanations, those produced by the vanilla PGExplainer, and those generated by our method.


\subsection{Efficiency Analysis} \label{sec:efficiency}

\noindent \textbf{Analysis of Iterations.} 
To determine whether only a small number of iterations is sufficient for stable performance gains, we vary the number of refinement iterations 
and evaluate the explanation results.
We report AUC-ROC results for iteration counts ranging from 0 to 5. Additionally, we implement a progressively shrinking top-$k$ strategy during iterative surrogate model training:
at each iteration, only the top-$k$ most important edges are retained in the explanation graph, with $k$ decreasing from 0.9 to 0.1 across iterations 1 through 5. The progressively shrinking strategy begins with broad subgraph exploration and incrementally focuses on discriminative structures. This process enforces finer attention to class-relevant structures while mitigating distractions from less informative edges.

\begin{wrapfigure}[7]{r}{0.48\textwidth}
    \centering    
    \vspace{-0.5cm}
    \includegraphics[width=1.0\linewidth]{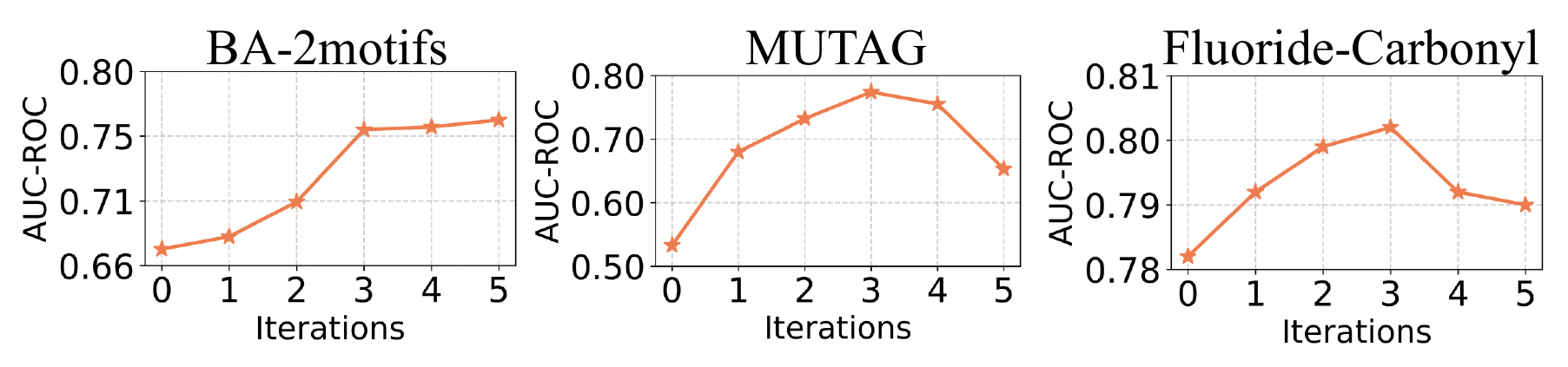}
    \caption{AUC-ROC performance with different numbers of iterations.}
    \label{fig:hyperparameter_n}
    \vspace{-5.0cm}
\end{wrapfigure}

As shown in Figure~\ref{fig:hyperparameter_n}, on {\bamo}, the AUC-ROC steadily increases during the first few iterations and plateaus after iteration 3, suggesting that iterative refinement and stronger substructure contrast effectively enhance explanation fidelity and model robustness. 
On {\mutag} and {\fluo}, performance peaks at iteration 3 and slightly declines thereafter. This suggests that overly aggressive pruning of edges, resulting from smaller $k$ values in later iterations, may remove informative substructures, thus impairing both explanation fidelity and model generalization.
Notably, a substantial improvement already appears at iteration 1 across all datasets compared with the original model at iteration 0, demonstrating that {\ours} does not rely on many refinement steps and showing that even one or two iterations are sufficient to achieve stable performance gains.

Moreover, the shrinking top-$k$ strategy encourages the explainer to focus on the most important edges over time. The observed improvements suggest that this gradual refinement enhances explanation accuracy and improves the robustness of the GNN to noise from less important structures. We also find that the performance after iterative training
consistently surpasses the original baseline across all iterations, which suggests the robustness of our method.
\begin{wrapfigure}{r}{0.42\linewidth}
    \centering   
    \includegraphics[width=0.76\linewidth]{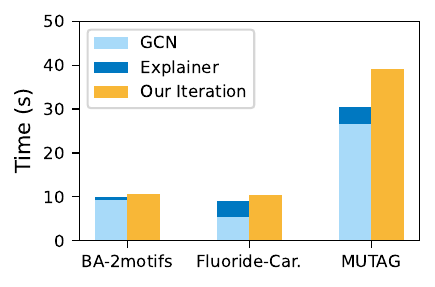}
    \caption{Average training time. GCN and Explainer jointly represent the standard pipeline.}
    \label{fig:time_cost}
    \vspace{-0.5cm}
\end{wrapfigure}

\noindent \textbf{Running Time.} \label{app:time} 
Figure \ref{fig:time_cost} illustrates the computational efficiency of our framework by reporting the average runtime calculated over 10 independent runs. We compare the standard approach, which includes GCN training and PGExplainer optimization, against one complete iteration of STORE that involves stochastic graph construction and surrogate model training. The results indicate that the additional overhead is modest across all datasets. Notably, for {\bamo} and {\fluo}, the runtime for one iteration is nearly equivalent to the standard training baseline, while the increase on {\mutag} remains within a highly practical range. Given the significant improvements in explanation accuracy shown in Figure \ref{fig:hyperparameter_n}, these results confirm that our iterative framework achieves substantial performance gains with minimal computational cost.


\subsection{Ablation Studies}
\noindent \textbf{Sampling Methods.} To validate our Gaussian-based re-weighting, we compare it against two baselines: 
{i) Random Perturbation: Edge weights are sampled uniformly from $(0,1)$, ignoring explainer importance. 
{ii) Uniform-Based Sampling: A structured uniform approach where explanatory edges are sampled from $W_1 \sim \text{Uniform}(u, 1)$, and non-explanatory edges are scaled by noise $Z \sim \text{Uniform}(0, 1)$ such that $W_2 = W_1 \cdot Z$. This concentrates non-explanatory weights in $(0, u)$, providing separation but lacking the smooth distributional properties of our Gaussian approach.


\begin{table}[ht]
\centering
\small
\caption{Performance comparison with different sampling methods}
\begin{tabular}{p{1.5cm}>{\centering\arraybackslash}m{2.0cm}>{\centering\arraybackslash}m{2.0cm}>{\centering\arraybackslash}m{2.0cm}}
\toprule
 & {\bamo} & {\mutag} & Fluoride-Car. \\
\midrule
Random   & 0.631$_{\pm 0.282}$ & 0.513$_{\pm 0.264}$ & 0.763$_{\pm 0.035}$ \\
Uniform  & 0.678$_{\pm 0.241}$ & 0.757$_{\pm 0.154}$ & 0.795$_{\pm 0.022}$ \\
Gaussian & 0.758$_{\pm 0.232}$ & 0.774$_{\pm 0.189}$ & 0.802$_{\pm 0.020}$ \\
\bottomrule
\end{tabular}
\label{tab:ablation_sampling}
\end{table}

Table~\ref{tab:ablation_sampling} reports the performance on {\bamo}, {\mutag}, and {\fluo} under different sampling strategies. We observe that the random perturbation yields the lowest AUC-ROC scores and the largest variances, indicating that random noise disrupts the learning process, leading to reduced stability. The Uniform-based strategy leads to notable performance gains compared to random perturbation, especially on {\mutag} and {\fluo}, demonstrating that incorporating edge importance helps preserve informative structures. The Gaussian-based strategy used in {\ours} achieves the best performance on all three datasets. It outperforms the other two variants in average performance, suggesting that it provides a better balance between edge importance preservation and controlled perturbation.

\noindent \textbf{Effects of $\Delta \mu$.}
We conduct a hyperparameter analysis to investigate how the difference $\Delta \mu$ between the truncated Gaussian distributions for explanatory and non-explanatory edges affects the explainer's performance. As shown in Figure~\ref{fig:hyperparameter_mu}, 
we vary $\Delta \mu$ from 0.1 to 0.7 and report the AUC-ROC scores on {\bamo}, {\mutag}, and {\fluo}.  
For the larger values $\Delta\mu=0.6$ and $\Delta\mu=0.7$, we reduce the boundary margin in the mean-sampling interval to keep it non-empty, while keeping the truncated Gaussian sampling on $[0,1]$.
\begin{wrapfigure}[7]{r}{0.48\textwidth}
    \centering  
    \vspace{-0.3cm}
    \includegraphics[width=1.0\linewidth]{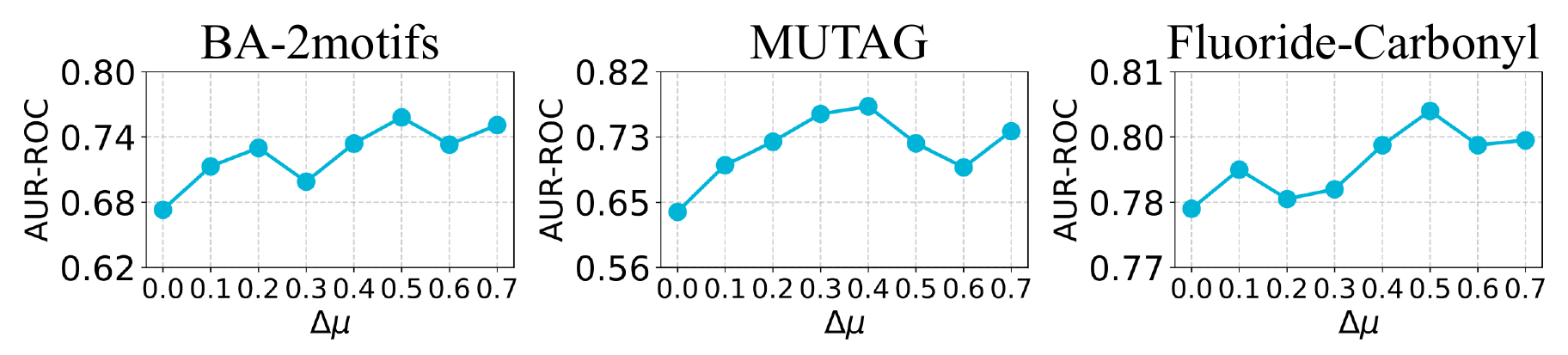}
    \caption{AUC-ROC performance with different $\Delta \mu$ for PGExplainer.}
    \label{fig:hyperparameter_mu}
\end{wrapfigure}
We observe that the performance remains relatively stable across a wide range of $\Delta \mu$ values. On {\bamo} and {\fluo}, the AUC-ROC improves steadily with increasing $\Delta \mu$ and reaches its peak around 0.5. On {\mutag}, the performance improves initially and peaks at $\Delta \mu = 0.4$, then drops slightly beyond that. These findings indicate that a moderate separation between important and unimportant edges helps the model better distinguish informative substructures. 

\subsection{Alleviating Distribution Shifts}
In this section, we visualize the classification accuracy under different edge weight configurations, where the x-axis denotes explanation edge weights and the y-axis denotes non-explanation edge weights. Figure~\ref{fig:heatmap} presents heat maps on {\bamo}, {\mutag}, and {\fluo}. For each dataset, we compare the original GCN model with our explanation-aware surrogate model. 

\begin{figure}[h]
    \centering
    \includegraphics[width=0.95\textwidth]{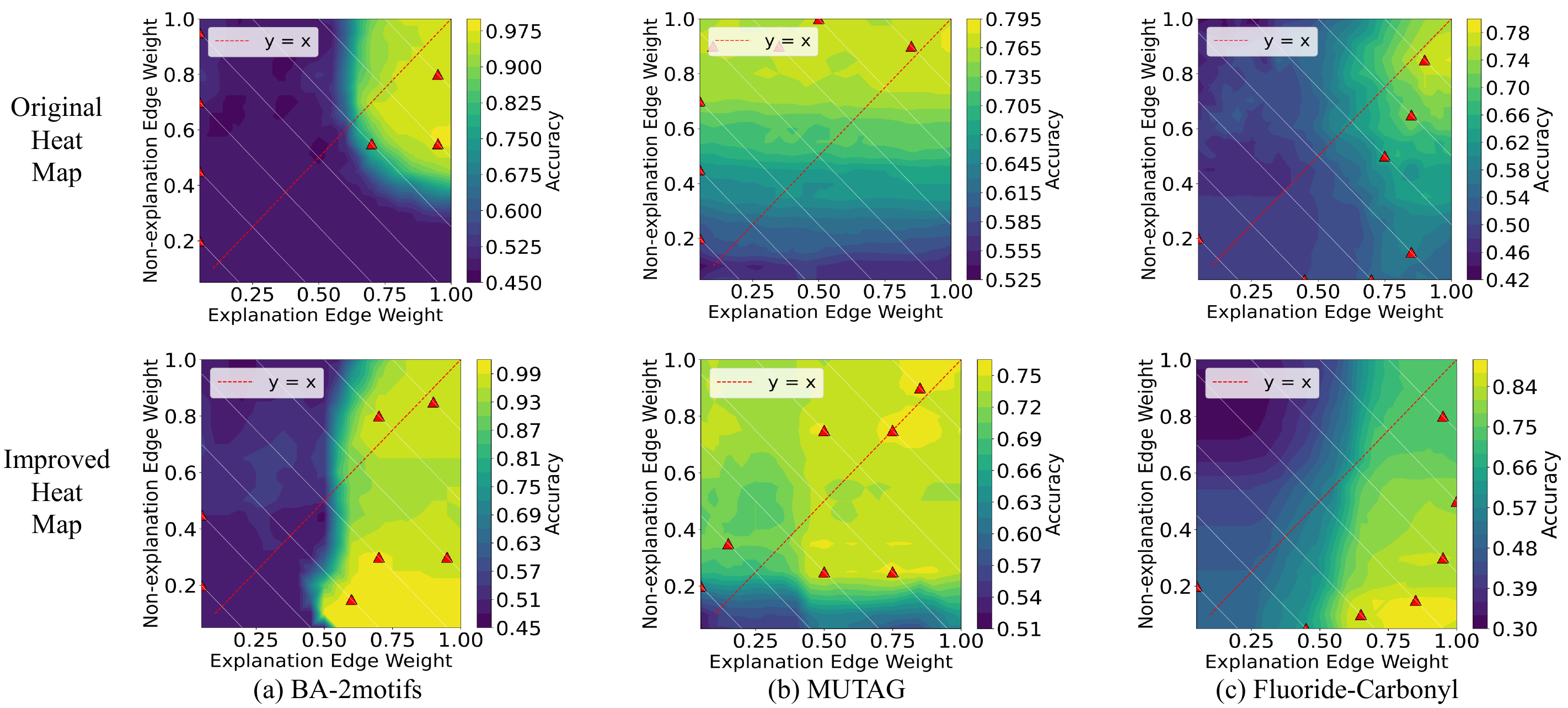}
    \caption{Heat maps of classification accuracy under different edge weights.}
    \label{fig:heatmap}
\end{figure}

In the original model, we observe that only the upper-right region achieves high accuracy, where both explanation and non-explanation edge weights are large. In contrast, accuracy drops significantly in the lower-right region, where the graph contains only explanation edges. These results indicate that the model heavily relies on the global graph structure and fails to generalize when presented with explanation subgraphs, revealing the OOD problem. On the {\mutag} dataset, we further find that the model can still achieve high accuracy even when explanation edge weights are small and non-explanation edge weights are large. This suggests that the model may exploit spurious background structures to make predictions, thereby undermining the faithfulness of explanation subgraphs. After applying explanation-aware iteration training, the model achieves consistently higher accuracy in a broader range of configurations, especially in the lower-right region. This indicates that the model has learned to better focus on explanation structures and that our method mitigates the OOD problem. In Appendix~\ref{app:validity}, we further report classification results and qualitative visualizations, confirming that the surrogate
GCN preserves both performance and explanatory consistency.

\section{Conclusion}
In this paper, we identify a critical flaw in post-hoc GNN explainability: the distributional shift caused by applying continuous explanation masks to models trained on unweighted graphs. To overcome this, we propose {\ours}, an iterative framework that constructs a robust surrogate model tailored to handle the weighted graphs inherent to the explanation process. By alternating between explanation identification and stochastic re-weighting, we ensure this surrogate serves as a faithful proxy for the original model while maintaining gradient reliability during mask optimization. Extensive experiments demonstrate that our method effectively mitigates the distribution mismatch, providing a general and model-agnostic solution for enhancing GNN interpretability.

\noindent\textbf{Limitations.} This work focuses on static graph classification. Extending the framework to node-level prediction and larger-scale graph learning, as well as developing adaptive stopping criteria for surrogate refinement, are promising directions for future work.



\begin{ack}\label{sec:acknowledgment}
This work was supported in part by the National Science Foundation under grants ECCS-2242700 and IIS-2529283, the Singapore Ministry of Education (MOE) Academic Research Fund (AcRF) Tier 1 Grant (26-SCIS-SMU-008), and the SMU Presidential Early Career Professorship (SMU-PEP) Start-Up Grant.

\end{ack}

\bibliographystyle{plainnat}
\bibliography{reference}

@inproceedings{zheng2024towards,
title={Towards Robust Fidelity for Evaluating Explainability of Graph Neural Networks},
author={Xu Zheng and Farhad Shirani and Tianchun Wang and Wei Cheng and Zhuomin Chen and Haifeng Chen and Hua Wei and Dongsheng Luo},
booktitle={The Twelfth International Conference on Learning Representations},
year={2024}
}

@article{hooker2019benchmark,
  title={A benchmark for interpretability methods in deep neural networks},
  author={Hooker, Sara and Erhan, Dumitru and Kindermans, Pieter-Jan and Kim, Been},
  journal={Advances in neural information processing systems},
  volume={32},
  year={2019}
}

@inproceedings{ribeiro2016should,
  title={" Why should i trust you?" Explaining the predictions of any classifier},
  author={Ribeiro, Marco Tulio and Singh, Sameer and Guestrin, Carlos},
  booktitle={Proceedings of the 22nd ACM SIGKDD international conference on knowledge discovery and data mining},
  pages={1135--1144},
  year={2016}
}

@article{amara2023ginx,
  title={GInX-Eval: Towards In-Distribution Evaluation of Graph Neural Network Explanations},
  author={Amara, Kenza and El-Assady, Mennatallah and Ying, Rex},
  journal={arXiv preprint arXiv:2309.16223},
  year={2023}
}

@article{luo2024towards,
  title={Towards inductive and efficient explanations for graph neural networks},
  author={Luo, Dongsheng and Zhao, Tianxiang and Cheng, Wei and Xu, Dongkuan and Han, Feng and Yu, Wenchao and Liu, Xiao and Chen, Haifeng and Zhang, Xiang},
  journal={IEEE Transactions on Pattern Analysis and Machine Intelligence},
  volume={46},
  number={8},
  pages={5245--5259},
  year={2024},
  publisher={IEEE}
}

@article{xu2018powerful,
  title={How powerful are graph neural networks?},
  author={Xu, Keyulu and Hu, Weihua and Leskovec, Jure and Jegelka, Stefanie},
  journal={ICLR},
  year={2019}
}

@article{ying2019gnnexplainer,
  title={Gnnexplainer: Generating explanations for graph neural networks},
  author={Ying, Zhitao and Bourgeois, Dylan and You, Jiaxuan and Zitnik, Marinka and Leskovec, Jure},
  journal={Advances in neural information processing systems},
  volume={32},
  year={2019}
}

@article{yuan2022explainability,
  title={Explainability in graph neural networks: A taxonomic survey},
  author={Yuan, Hao and Yu, Haiyang and Gui, Shurui and Ji, Shuiwang},
  journal={IEEE transactions on pattern analysis and machine intelligence},
  volume={45},
  number={5},
  pages={5782--5799},
  year={2022},
  publisher={IEEE}
}

@inproceedings{zhang2023mixupexplainer,
  title={Mixupexplainer: Generalizing explanations for graph neural networks with data augmentation},
  author={Zhang, Jiaxing and Luo, Dongsheng and Wei, Hua},
  booktitle={Proceedings of the 29th ACM SIGKDD Conference on Knowledge Discovery and Data Mining},
  pages={3286--3296},
  year={2023}
}

@inproceedings{chen2024generating,
  title={Generating In-Distribution Proxy Graphs for Explaining Graph Neural Networks},
  author={Chen, Zhuomin and Zhang, Jiaxing and Ni, Jingchao and Li, Xiaoting and Bian, Yuchen and Islam, Md Mezbahul and Mondal, Ananda and Wei, Hua and Luo, Dongsheng},
  booktitle={Forty-first International Conference on Machine Learning},
  year={2024}
}

@inproceedings{huang2024factorized,
  title={Factorized explainer for graph neural networks},
  author={Huang, Rundong and Shirani, Farhad and Luo, Dongsheng},
  booktitle={Proceedings of the AAAI conference on artificial intelligence},
  volume={38},
  pages={12626--12634},
  year={2024}
}

@article{luo2020parameterized,
  title={Parameterized explainer for graph neural network},
  author={Luo, Dongsheng and Cheng, Wei and Xu, Dongkuan and Yu, Wenchao and Zong, Bo and Chen, Haifeng and Zhang, Xiang},
  journal={Advances in neural information processing systems},
  volume={33},
  pages={19620--19631},
  year={2020}
}

@article{agarwal2023evaluating,
  title={Evaluating explainability for graph neural networks},
  author={Agarwal, Chirag and Queen, Owen and Lakkaraju, Himabindu and Zitnik, Marinka},
  journal={Scientific Data},
  volume={10},
  number={1},
  pages={144},
  year={2023},
  publisher={Nature Publishing Group UK London}
}

@inproceedings{yuan2020xgnn,
  title={Xgnn: Towards model-level explanations of graph neural networks},
  author={Yuan, Hao and Tang, Jiliang and Hu, Xia and Ji, Shuiwang},
  booktitle={Proceedings of the 26th ACM SIGKDD international conference on knowledge discovery \& data mining},
  pages={430--438},
  year={2020}
}

@inproceedings{yuan2021explainability,
  title={On explainability of graph neural networks via subgraph explorations},
  author={Yuan, Hao and Yu, Haiyang and Wang, Jie and Li, Kang and Ji, Shuiwang},
  booktitle={International conference on machine learning},
  pages={12241--12252},
  year={2021},
  organization={PMLR}
}

@inproceedings{wang2022gnninterpreter,
  author       = {Xiaoqi Wang and
                  Han{-}Wei Shen},
  title        = {GNNInterpreter: {A} Probabilistic Generative Model-Level Explanation
                  for Graph Neural Networks},
  booktitle    = {The Eleventh International Conference on Learning Representations,
                  {ICLR} 2023, Kigali, Rwanda, May 1-5, 2023},
  year         = {2023}
}

@article{wang2021towards,
  title={Towards multi-grained explainability for graph neural networks},
  author={Wang, Xiang and Wu, Yingxin and Zhang, An and He, Xiangnan and Chua, Tat-Seng},
  journal={Advances in Neural Information Processing Systems},
  volume={34},
  pages={18446--18458},
  year={2021}
}

@article{vu2020pgm,
  title={Pgm-explainer: Probabilistic graphical model explanations for graph neural networks},
  author={Vu, Minh and Thai, My T},
  journal={Advances in neural information processing systems},
  volume={33},
  pages={12225--12235},
  year={2020}
}

@inproceedings{kipf2017semisupervised,
    title={Semi-Supervised Classification with Graph Convolutional Networks},
    author={Thomas N. Kipf and Max Welling},
    booktitle={International Conference on Learning Representations},
    year={2017}
}

@article{kingma2014adam,
  title={Adam: A method for stochastic optimization},
  author={Kingma, Diederik P and Ba, Jimmy},
  journal={arXiv preprint arXiv:1412.6980},
  year={2014}
}

@inproceedings{zhang2022gstarx,
 author = {Zhang, Shichang and Liu, Yozen and Shah, Neil and Sun, Yizhou},
 booktitle = {Advances in Neural Information Processing Systems},
 pages = {19810--19823},
 title = {GStarX: Explaining Graph Neural Networks with Structure-Aware Cooperative Games},
 volume = {35},
 year = {2022}
}

@inproceedings{li2023dag,
title={{DAG} Matters! {GF}lowNets Enhanced Explainer for Graph Neural Networks},
author={Wenqian Li and Yinchuan Li and Zhigang Li and Jianye HAO and Yan Pang},
booktitle={The Eleventh International Conference on Learning Representations },
year={2023}
}

@article{wu2018moleculenet,
  title={MoleculeNet: a benchmark for molecular machine learning},
  author={Wu, Zhenqin and Ramsundar, Bharath and Feinberg, Evan N and Gomes, Joseph and Geniesse, Caleb and Pappu, Aneesh S and Leswing, Karl and Pande, Vijay},
  journal={Chemical science},
  volume={9},
  number={2},
  pages={513--530},
  year={2018},
  publisher={Royal Society of Chemistry}
}

@inproceedings{wang2019knowledge,
  title={Knowledge-aware graph neural networks with label smoothness regularization for recommender systems},
  author={Wang, Hongwei and Zhang, Fuzheng and Zhang, Mengdi and Leskovec, Jure and Zhao, Miao and Li, Wenjie and Wang, Zhongyuan},
  booktitle={Proceedings of the 25th ACM SIGKDD international conference on knowledge discovery \& data mining},
  pages={968--977},
  year={2019}
}

@inproceedings{bu2023improving,
  title={On Improving the Cohesiveness of Graphs by Merging Nodes: Formulation, Analysis, and Algorithms},
  author={Bu, Fanchen and Shin, Kijung},
  booktitle={Proceedings of the 29th ACM SIGKDD Conference on Knowledge Discovery and Data Mining},
  pages={117--129},
  year={2023}
}

@inproceedings{anklin2021learning,
  title={Learning whole-slide segmentation from inexact and incomplete labels using tissue graphs},
  author={Anklin, Valentin and Pati, Pushpak and Jaume, Guillaume and Bozorgtabar, Behzad and Foncubierta-Rodriguez, Antonio and Thiran, Jean-Philippe and Sibony, Mathilde and Gabrani, Maria and Goksel, Orcun},
  booktitle={Medical Image Computing and Computer Assisted Intervention--MICCAI 2021: 24th International Conference, Strasbourg, France, September 27--October 1, 2021, Proceedings, Part II 24},
  pages={636--646},
  year={2021},
  organization={Springer}
}

@article{rai2020explainable,
  title={Explainable AI: From black box to glass box},
  author={Rai, Arun},
  journal={Journal of the academy of marketing science},
  volume={48},
  pages={137--141},
  year={2020},
  publisher={Springer}
}

@article{albert2002statistical,
  title={Statistical mechanics of complex networks},
  author={Albert, R{\'e}ka and Barab{\'a}si, Albert-L{\'a}szl{\'o}},
  journal={Reviews of modern physics},
  volume={74},
  number={1},
  pages={47},
  year={2002},
  publisher={APS}
}

@article{dobson2003distinguishing,
  title={Distinguishing enzyme structures from non-enzymes without alignments},
  author={Dobson, Paul D and Doig, Andrew J},
  journal={Journal of molecular biology},
  volume={330},
  number={4},
  pages={771--783},
  year={2003},
  publisher={Elsevier}
}

@article{chen2026addressing,
  title={Addressing Structural Distribution Shift in Explanations for Graph Neural Networks},
  author={Chen, Zhuomin and Salehi, Hojat Allah and Schafir, Esteban and Zheng, Xu and Zhang, Jiaxing and Wei, Hua and Ni, Jingchao and Shirani, Farhad and Luo, Dongsheng},
  journal={IEEE transactions on pattern analysis and machine intelligence},
  year={2026},
  publisher={IEEE}
}

@inproceedings{lu2024goat,
  author       = {Shengyao Lu and
                  Keith G. Mills and
                  Jiao He and
                  Bang Liu and
                  Di Niu},
  title        = {GOAt: Explaining Graph Neural Networks via Graph Output Attribution},
  booktitle    = {The Twelfth International Conference on Learning Representations, {ICLR}},
  year         = {2024}
}

@inproceedings{lu2024eigsearch,
title={EiG-Search: Generating Edge-Induced Subgraphs for {GNN} Explanation in Linear Time},
author={Shengyao Lu and Bang Liu and Keith G. Mills and Jiao He and Di Niu},
booktitle={Forty-first International Conference on Machine Learning},
year={2024}
}

@InProceedings{pmlr-v235-bui24b,
  title = 	 {Explaining Graph Neural Networks via Structure-aware Interaction Index},
  author =       {Bui, Ngoc and Nguyen, Hieu Trung and Nguyen, Viet Anh and Ying, Rex},
  booktitle = 	 {Proceedings of the 41st International Conference on Machine Learning},
  pages = 	 {4854--4883},
  year = 	 {2024},
  volume = 	 {235},
  series = 	 {Proceedings of Machine Learning Research},
  publisher =    {PMLR}
}

@InProceedings{pmlr-v267-qu25a,
  title = 	 {{RISE}: Radius of Influence based Subgraph Extraction for 3{D} Molecular Graph Explanation},
  author =       {Qu, Jingxiang and Gao, Wenhan and Zhang, Jiaxing and Liu, Xufeng and Wei, Hua and Ling, Haibin and Liu, Yi},
  booktitle = 	 {Proceedings of the 42nd International Conference on Machine Learning},
  pages = 	 {50744--50761},
  year = 	 {2025},
  volume = 	 {267},
  series = 	 {Proceedings of Machine Learning Research},
  publisher =    {PMLR}
}

@inproceedings{yang2025from,
title={From {GNN}s to Trees: Multi-Granular Interpretability for Graph Neural Networks},
author={Jie Yang and Yuwen Wang and Kaixuan Chen and Tongya Zheng and Yihe Zhou and Zhenbang Xiao and Ji Cao and Mingli Song and Shunyu Liu},
booktitle={The Thirteenth International Conference on Learning Representations},
year={2025}
}

@article{zhang2024regexplainer,
  title={Regexplainer: Generating explanations for graph neural networks in regression tasks},
  author={Zhang, Jiaxing and Chen, Zhuomin and Mei, Hao and Da, Longchao and Luo, Dongsheng and Wei, Hua},
  journal={Advances in Neural Information Processing Systems},
  volume={37},
  pages={79282--79306},
  year={2024}
}

@article{chen2023d4explainer,
  title={D4explainer: In-distribution explanations of graph neural network via discrete denoising diffusion},
  author={Chen, Jialin and Wu, Shirley and Gupta, Abhijit and Ying, Rex},
  journal={Advances in Neural Information Processing Systems},
  volume={36},
  pages={78964--78986},
  year={2023}
}

@article{saha2025gen,
  author       = {Sayan Saha and
                  Monidipa Das and
                  Sanghamitra Bandyopadhyay},
  title        = {Gen-GraphEx: Generative In-Distribution Graph Explanations for Time-Efficient Model-Level Interpretability of GNNs},
  journal      = {{IEEE} Transactions on Neural Networks and Learning Systems},
  volume       = {36},
  number       = {11},
  pages        = {19775--19789},
  year         = {2025}
}

\newpage

\appendix
\onecolumn
\appendix
\section{Theoretical Analysis: Full Proofs}
\label{app:theory}

This appendix provides the detailed definitions and proofs omitted from the main text. We follow the formalism of \citet{zheng2024towards}.

\subsection{Preliminaries}

\begin{definition}[Explainable Classification Task {\citep{zheng2024towards}}]
\label{def:exp_prob}
A task with distribution $P_{Y,\mX,\mathbf{A}}$ is $(s,\kappa)$-explainable if there exists an explanation function $\Psi:{\mathcal{G}}\to 2^{\mathcal{V}}\times 2^{\mathcal{E}}$ such that
\[
I\!\left(Y;{G}\,\middle|\,\mathds{1}_{\Psi({G})}\right)\le \kappa
\quad\text{and}\quad
\mathbb{E}_G\big[|\mathcal{E}_{\mathrm{exp}}|\big]\le s,
\]
where $\Psi({G})=(\mathcal{V}_{\mathrm{exp}},\mathcal{E}_{\mathrm{exp}})$.
\end{definition}

\begin{definition}[Explainable Classifier {\citep{zheng2024towards}}]
\label{def:exp_class}
For a classifier $f:{\mathcal{G}}\to\Delta_{\mathcal{Y}}$, let $\widehat{Y}\sim f({G})$. Then $f$ is $(s,\zeta)$-explainable if there exists $\Psi$ with
\[
I\!\left(\widehat{Y};{G}\,\middle|\,\mathds{1}_{\Psi({G})}\right)\le \zeta
\quad\text{and}\quad
\mathbb{E}_G\big[|\mathcal{E}_{\mathrm{exp}}|\big]\le s.
\]
\end{definition}

\paragraph{Condition 1 (Stability under supergraphs).}
For all ${g},{g}'$, if $\Psi({g})\subseteq {g}'$, then $\Psi({g}')=\Psi({g})$. This condition implies
\[
I(\widehat{Y};{G}\,|\,\mathds{1}_{\Psi({G})})
= I(\widehat{Y};{G}\,|\,\Psi({G})).
\]

\begin{definition}[Uniqueness of Task Explanation]
\label{def:unique}
A binary task is \emph{uniquely $(s,\kappa)$-explainable} if there exists an explanation $\Psi^\star$ satisfying Definition~\ref{def:exp_prob} and Condition~1, and for any other valid $\Psi$ we have
\[
\mathbb{P}\big(\Psi({G})=\Psi^\star({G})\big)=1.
\]
\end{definition}

\subsection{Proof of Proposition~\ref{prop:storeshared}}

\begin{proof}
The Theorem 1 in \citet{zheng2024towards} states that, if a task is $(s,\kappa)$-explainable with an explanation function satisfying Condition 1, then any classifier $f$ with error at most $\epsilon^\star+\delta$ is $(s,\eta)$-explainable for some $\eta\triangleq h_b(\tau)+\tau$, where $h_b(\tau)\triangleq-\tau\log{\tau}-(1-\tau)\log{1-\tau}$, $\tau\in [0,1]$ is the binary entropy function,
\begin{align*}
   \tau\triangleq \frac{\left( 2\sqrt{2}\epsilon^*+\sqrt{\xi} \right)^2}{2},
    \quad \xi \triangleq \delta+e_{max}(\epsilon^*,\kappa)-\epsilon^*,
    \quad \delta\in (0,9\epsilon^*-e_{max}(\epsilon^*,\kappa)), 
\end{align*} 
and $e_{\max}(\epsilon^\star,\kappa)$ is the error bound from the reverse-Fano inequality, defined in the Proposition 1 of \citet{zheng2024towards}. In particular, $\eta \to 0$ as $(\delta,\epsilon^\star,\kappa)\to 0$. The classifiers $f^{(0)}$ and $f^{(L)}$ defined in our paper satisfy the accuracy conditions with error at most $\epsilon^\star+\delta$. Thus, both $f^{(0)}$ and $f^{(L)}$ are $(s,\eta_i)$-explainable with respect to certain explanation functions.

Furthermore, the task admits a unique explanation $\Psi^\star$ by using the fact that the explanation $\Psi^\star$ satisfies Definition~\ref{def:exp_prob} and Condition~1 in Definition~\ref{def:unique}.
Thus, any valid explanation $\Psi$ that makes a classifier $(s,\eta)$-explainable must coincide with $\Psi^\star$ almost surely. Hence, both $f^{(0)}$ and $f^{(L)}$ are explained by the same $\Psi^\star$. This completes the proof.
\end{proof}






\subsection{Discussion}

The proof highlights two essential assumptions:
\begin{enumerate}
\item Condition 1, which allows conditioning on $\Psi({G})$ directly.
\item Uniqueness of the task explanation, which ensures both classifiers align with the same $\Psi^\star$.
\end{enumerate}
These strengthen the original statement and remove the gap caused by the possibility of multiple valid explanations.

\clearpage
\section{Algorithm} \label{app:algorithm}

\begin{algorithm}[h!]
\caption{Iterative Explanation Framework}
\begin{flushleft}
\textbf{Input:} A set of original graphs $\mathcal{G}$, GNN model $f(\cdot)$, explainer $\Psi(\cdot)$, number of iterations $L$ \\
\textbf{Output:} explainer $\Psi^{(L)}(\cdot)$
\end{flushleft}
\begin{algorithmic}[1]
\State $f^{(0)} \gets f$
\State $\Psi^{(0)} \gets \text{TrainExplainer}(f^{(0)}, \mathcal{G})$
\For{$l = 1$ to $L$}
    \State $\mathcal{G}^{\text{exp}} \gets \Psi^{(l-1)}(\mathcal{G})$ \Comment{Generate explanation graphs}
    \State $\mathcal{G}' \gets \text{AugmentGraph}(\mathcal{G}, \mathcal{G}^{\text{exp}})$ \Comment{Construct augmentation}
    \State $f^{(l)} \gets \text{TrainGNN}(\mathcal{G}')$ \Comment{Train surrogate GNN on augmented data} 
    \State $\Psi^{(l)} \gets \text{TrainExplainer}(f^{(l)}, \mathcal{G})$ \Comment{Retrain explainer on original graphs} 
\EndFor
\State \textbf{return} $\Psi^{(L)}(\cdot)$
\end{algorithmic}
\label{alg:process}
\end{algorithm}

\begin{algorithm}[h]
\caption{Construction of Augmented Graphs}
\begin{flushleft}
\textbf{Input:} A set of original graphs $\mathcal{G}$, corresponding explanation graphs $\mathcal{G}^{\text{exp}}$, mean shift $\Delta \mu$, top-$k$ edges to select, separation probability $\alpha$ \\
\textbf{Output:} A set of augmented graphs $\mathcal{G}'$
\end{flushleft}
\begin{algorithmic}[1]
\Statex Compute variance: $\sigma^2 = \frac{(\Delta \mu)^2}{2 (\Phi^{-1}(1-\alpha))^2}$
\Statex Sample $\mu_2 \sim \mathcal{U}(2\sigma, 1 - \Delta\mu - 2\sigma)$, set $\mu_1 = \mu_2 + \Delta \mu$
\State Initialize augmented graph set: $\mathcal{G}' \leftarrow \emptyset$
\For{$i = 1$ to $|\mathcal{G}|$}
    \State Copy original graph: $\mathcal{G}^{\text{aug},i} \leftarrow \mathcal{G}^{i}$
    \State Select top-$k$ explanation edges: $\mathcal{E}^{\text{exp}} \leftarrow \text{top-}k(\mathcal{G}^{\text{exp},i})$
    
    \For{each edge $e \in \mathcal{E}^{\text{exp}}$}
        \State Sample $W_{1,e} \sim \mathcal{TN}(\mu_1, \sigma^2; 0, 1)$
        \State Assign $w(e) \leftarrow W_{1,e}$ in $\mathcal{G}^{\text{aug},i}$
    \EndFor
    
    \For{each edge $e \in \mathcal{G}^{i} \setminus \mathcal{E}^{\text{exp}}$}
        \State Sample $W_{2,e} \sim \mathcal{TN}(\mu_2, \sigma^2; 0, 1)$
        \State Assign $w(e) \leftarrow W_{2,e}$ in $\mathcal{G}^{\text{aug},i}$
    \EndFor

    \State Add to augmented set: $\mathcal{G}' \leftarrow \mathcal{G}' \cup \{\mathcal{G}^{\text{aug},i}\}$
\EndFor
\State \textbf{return} $\mathcal{G}'$
\end{algorithmic}
\label{alg:aug_graphs}
\end{algorithm}

\clearpage
\section{Detailed Experimental Setup}

\subsection{Datasets} \label{app:data}
\begin{itemize} [leftmargin=*]
\item \textbf{{\bamo}} \citep{luo2020parameterized}. This synthetic dataset contains 1,000 graphs generated from the Barabási–Albert (BA) model. Each graph belongs to one of two classes: one with house-shaped motifs and the other with five-node cycle structures.  \\
\item \textbf{{\mutag}} \citep{luo2020parameterized}.
{\mutag} consists of 4,337 molecular graphs, each labeled by its mutagenic effect on the Gram-negative bacterium Salmonella typhimurium. \\
\item \textbf{{\benz}} \citep{agarwal2023evaluating}.
{\benz} consists of 12,000 molecular graphs, categorized into two groups—those containing benzene rings and those without. \\
\item \textbf{{\alk}} \citep{agarwal2023evaluating}.
This dataset contains 4,326 molecules. 
For positive labels, we consider molecules that have both alkane and carbonyl functional groups in one molecule. \\
\item \textbf{{\fluo}} \citep{agarwal2023evaluating}. 
{\fluo} consists of 8,671 molecular graphs, whose ground-truth explanations are based on both the presence of fluoride atoms and carbonyl functional groups in the molecule. \\
\item \textbf{D\&D} \citep{dobson2003distinguishing}. D\&D consists of protein graphs labeled as either enzymes or non-enzymes. Nodes correspond to amino acids, and an edge is added between two nodes whenever their spatial distance is no greater than 6 Angstroms. \\
\item \textbf{Graph-SST2} \citep{yuan2022explainability}. This dataset is a sentiment-analysis benchmark in which each sentence is converted into a graph representation, and the resulting graphs are annotated with their corresponding sentiment labels. \\

\end{itemize}
A summary of dataset statistics is provided in Table~\ref{tab:dataset}. 


\begin{table*}[ht]
\centering
\caption{Statistics of datasets used for graph classification task}
\begin{tabular}{lccccc}
\hline
Dataset  & Domain & \#Graphs & Avg.\#nodes & Avg.\#edges  & \#Feature  \\
\hline
{\bamo}    & Synthetic   & 1,000            & 25              & 51               & 10                         \\
{\mutag}      &  Biochemical molecules      & 4,337            & 30.32           & 30.77      & 14               \\
{\benz}     & Biochemical molecules      & 12,000           &  20.58            & 43.65  & 14               \\
Alkane-Car.     &  Biochemical molecules      & 4,326            & 21.13            & 44.95   & 14              \\
Fluoride-Car.   &  Biochemical molecules     & 8,671            & 21.36            & 45.37                            & 14               \\
D\&D      & Bioinformatics    & 1,178              & 284.32        & 715.66    & 89   \\
Graph-SST2 & Natural language &  70,042 & 10.20 & 18.40 & 768  \\
\hline
\end{tabular}
\label{tab:dataset}
\end{table*}

\subsection{Baselines} \label{app:baseline}

\begin{itemize} [leftmargin=*]
\item \textbf{GNNExplainer} \citep{ying2019gnnexplainer}.
GNNExplainer learns edge and node feature masks to identify graph components relevant to a prediction. It applies these masks element-wise to the input graph and optimizes them through a mutual-information objective. \\
\item \textbf{PGExplainer} \citep{luo2020parameterized}.
Building on GNNExplainer, PGExplainer models edge selection using Bernoulli distributions under a Gilbert-graph assumption. The inclusion probability of each edge is predicted from the concatenated embeddings of its endpoint nodes, and Gumbel-Softmax enables differentiable sampling during training. \\
\item \textbf{ReFine} \citep{wang2021towards}.
ReFine adopts a two-stage approach. It first estimates edge selection probabilities at the class level using mutual-information and contrastive objectives. These probabilities then guide edge sampling during instance-level fine-tuning to obtain explanations that retain information relevant to the model's prediction. \\
\item \textbf{MixupExplainer} \citep{zhang2023mixupexplainer}. 
MixupExplainer uses a non-parametric augmentation strategy to reduce distribution shifts. It combines the selected explanation with a background subgraph sampled from another graph without conditioning on its class label. \\
\item \textbf{ProxyExplainer} \citep{chen2024generating}. 
ProxyExplainer mitigates the distributional shift in explanation by generating in-distribution proxy graphs via a parametric graph generator, ensuring both fidelity and alignment with the original graphs. 
\end{itemize}

\subsection{Implementation} \label{app:implementation}
We use the Adam optimizer~\citep{kingma2014adam} with a learning rate of $1 \times 10^{-3}$ for training the GNN, and use the same optimizer with a weight decay of $5 \times 10^{-4}$ for training the explainer. 
We use a linearly decreasing top-schedule, with $k$ = {0.9,0.7,0.5} for iterations 1–3, respectively. At each iteration, we retain the top-$k$ fraction of edges.
The AUC-ROC metric is adopted for quantitative evaluation. All experiments are conducted on a Linux machine equipped with four NVIDIA A100-PCIE GPUs, each with 40 GB of memory. The CUDA version is 12.3, and the driver version is 545.23.08.

\section{Extra Experimental Study} 

\subsection{Experiments with GIN Model} \label{app:gin}

In Table~\ref{tab:exp_gin}, we report results using GIN as the backbone architecture. The results show that {\ours} consistently improves the performance of various explanation methods across most datasets. For example, PGExplainer achieves a 29.2\% improvement on {\benz} and a 7.7\% gain on {\fluo}, while ReFine and ProxyExplainer also show improvements on {\bamo} and {\fluo}, respectively. These results further demonstrate the model-agnostic nature of our framework: even when switching to a different GNN architecture, the iterative refinement
process still improves explanation quality.

\begin{table*}[ht]
    \centering
    \caption{Performance comparison of different explanation methods using GIN as the backbone.}
    \begin{tabular}{m{2.5cm}|>{\centering\arraybackslash}m{1.8cm}>{\centering\arraybackslash}m{1.8cm}>{\centering\arraybackslash}m{1.8cm}>{\centering\arraybackslash}m{1.8cm}>{\centering\arraybackslash}m{1.8cm}}
    \hline
    Explainer Method & {\bamo} & {\mutag} & {\benz} & Alkane-Car.  & Fluoride-Car.   \\ \hline
    GNNExplainer   & 0.511$_{\pm 0.005}$ & 0.553$_{\pm 0.127}$ & 0.469$_{\pm 0.079}$ & 0.502$_{\pm 0.126}$  & 0.501$_{\pm 0.029}$  \\ 
                           {\scriptsize \hfill + \ours}    & \textbf{0.544$_{\pm 0.053}$} & \textbf{0.589$_{\pm 0.114}$} & \textbf{0.514$_{\pm 0.018}$} & \textbf{0.552$_{\pm 0.071}$} & \textbf{0.513$_{\pm 0.005}$} \\ 
    \hline
    PGExplainer & 0.756$_{\pm 0.315}$ & 0.489$_{\pm 0.246}$ & 0.617$_{\pm 0.260}$ & 0.495$_{\pm 0.259}$ & 0.678$_{\pm 0.088}$  \\ 
                           {\scriptsize \hfill + \ours}  & \textbf{0.791$_{\pm 0.176}$}  & \textbf{0.657$_{\pm 0.209}$} & \textbf{0.797$_{\pm 0.108}$} & \textbf{0.631$_{\pm 0.258}$} & \textbf{0.730$_{\pm 0.118}$}  \\ 
    \hline
    Refine & 0.518$_{\pm 0.066}$ & 0.535$_{\pm 0.225}$ & 0.499$_{\pm 0.127}$ & 0.603$_{\pm 0.190}$ & 0.551$_{\pm 0.074}$  \\ 
                           {\scriptsize \hfill + \ours}  & \textbf{0.584$_{\pm 0.066}$}   & \textbf{0.537$_{\pm 0.150}$} & \textbf{0.638$_{\pm 0.158}$} & \textbf{0.650$_{\pm 0.117}$} & \textbf{0.597$_{\pm 0.054}$} \\ 
    \hline
    MixupExplainer & 0.738$_{\pm 0.281}$ & \textbf{0.480$_{\pm 0.178}$} & 0.593$_{\pm 0.227}$ & 0.420$_{\pm 0.260}$ & 0.767$_{\pm 0.088}$  \\ 
                           {\scriptsize \hfill + \ours}   & \textbf{0.759$_{\pm 0.161}$} & 0.475$_{\pm 0.192}$ & \textbf{0.649$_{\pm 0.213}$} & \textbf{0.464$_{\pm 0.236}$} & \textbf{0.785$_{\pm 0.065}$}  \\ 
    \hline
    ProxyExplainer & 0.621$_{\pm 0.205}$ & 0.567$_{\pm 0.110}$ & 0.537$_{\pm 0.139}$ & 0.465$_{\pm 0.156}$ & \textbf{0.605$_{\pm 0.055}$}  \\ 
                           {\scriptsize \hfill + \ours}  & \textbf{0.659$_{\pm 0.173}$}  & \textbf{0.644$_{\pm 0.131}$} & \textbf{0.582$_{\pm 0.136}$} & \textbf{0.582$_{\pm 0.129}$} & 0.600$_{\pm 0.067}$  \\ 
    \hline
    \end{tabular}
    \label{tab:exp_gin}
\end{table*}

\subsection{Validity of Robust Surrogate Model}
\label{app:validity}

In Section~\ref{sec:exp:quantitative}, we have demonstrated the effectiveness of {\ours} in generating high-quality explanations. These results confirm the practical utility of addressing the distributional shift. To further validate that the robust surrogate remains a faithful proxy for the original model, we directly compare their behaviors on the original unweighted graphs. Specifically, we evaluate whether the surrogate preserves the original model's predictive performance, sample-wise predictions, and explanatory structures.

\noindent \textbf{Predictive Consistency.}
\label{app:iterationacc}
First, we examine whether the robust surrogate preserves the predictive capability of the original model. Table \ref{tab:pretrainresults} reports the graph classification performance of the original GCN versus the final robust surrogate on the original unweighted test set. For {\mutag}, we observe slight improvements in all metrics, while for {\bamo} and {\fluo}, the performance remains highly stable. These findings indicate that while the surrogate model extends its domain to handle weighted graphs, it retains the predictive performance of the original model.

Since similar accuracy alone does not guarantee sample-wise consistency, we further evaluate the direct agreement between the two models in Table~\ref{tab:surrogate_agreement}. Let $\hat y_i^{(0)}$ and $\hat y_i^{(L)}$ denote the predictions of the original and surrogate GCNs, respectively. We define prediction agreement as the fraction of test samples satisfying $\hat y_i^{(0)}=\hat y_i^{(L)}$. We also report the IoU between the correctly predicted sample sets, where $C^{(0)}=\{i:\hat y_i^{(0)}=y_i\}$ and $C^{(L)}=\{i:\hat y_i^{(L)}=y_i\}$. For reference, we include a simple independence baseline computed from the two models' marginal test accuracies. The consistently high agreement suggests that the surrogate largely preserves the decision behavior of the original GNN rather than merely matching its overall accuracy.

\begin{table}[ht]
\centering
\caption{Graph classification performance of the original GCN model and the final robust surrogate GCN model.}
\label{tab:pretrainresults}
\begin{tabular}{ll>{\centering\arraybackslash}m{2.7cm}>{\centering\arraybackslash}m{2.7cm}>{\centering\arraybackslash}m{2.7cm}}
\toprule
 & & {\bamo} & {\mutag} & {\fluo} \\
\midrule
\multirow{2}{*}{Train Acc} & Original     & 0.923$_{\pm 0.061}$   & 0.787$_{\pm 0.026}$   & 0.764$_{\pm0.036}$          \\
      & {Surrogate}     & 0.915$_{\pm 0.057}$   & 0.802$_{\pm 0.010}$    & 0.757$_{\pm 0.038}$  \\
\hline
\multirow{2}{*}{Val Acc} & Original     & 0.942$_{\pm 0.050}$     & 0.785$_{\pm 0.025}$   & 0.773$_{\pm 0.044}$     \\
      & {Surrogate}     & 0.933$_{\pm 0.062}$     & 0.791$_{\pm 0.011}$       & 0.766$_{\pm 0.033}$   \\
\hline
\multirow{2}{*}{Test Acc} & Original     & 0.934$_{\pm 0.049}$    & 0.771$_{\pm 0.026}$    & 0.750$_{\pm 0.053}$     \\
      & {Surrogate}     & 0.926$_{\pm 0.062}$    & 0.786$_{\pm 0.014}$    & 0.752$_{\pm 0.040}$   \\
\bottomrule
\end{tabular}
\end{table}

\begin{table}[ht]
\centering
\caption{Prediction and explanation agreement between the original GCN and the final robust surrogate GCN.}
\label{tab:surrogate_agreement}
\begin{tabular}{p{2.7cm}>{\centering\arraybackslash}m{2.7cm}>{\centering\arraybackslash}m{2.7cm}>{\centering\arraybackslash}m{2.7cm}}
\toprule
Metric & {\bamo} & {\mutag} & {\fluo} \\
\midrule
Indep. baseline & 87.0\% & 65.5\% & 62.6\% \\
Pred. agreement & 93.5\% & 80.3\% & 82.8\%\\
Correct-set IoU & 92.6\% & 78.1\% & 79.5\%\\
\midrule
Exp. overlap & 89.7\% & 76.8\% & 84.4\%\\
Exp. IoU & 81.9\% & 72.6\% & 76.3\%\\
\bottomrule
\end{tabular}
\end{table}

\noindent \textbf{Consistency of Explanation Structures.} 
\label{app:preserve}
We further evaluate whether the surrogate model preserves the explanatory structures of the original model. For each test graph, we extract the top-$k$ edges according to the explanation scores from the original and surrogate explainers, denoted by $E_i^{(0)}$ and $E_i^{(L)}$, respectively. Table~\ref{tab:surrogate_agreement} reports the average explanation overlap, $|E_i^{(0)}\cap E_i^{(L)}|/\min\{|E_i^{(0)}|,|E_i^{(L)}|\}$, and the average explanation IoU, $|E_i^{(0)}\cap E_i^{(L)}|/|E_i^{(0)}\cup E_i^{(L)}|$. The results show substantial explanation-level agreement across datasets, suggesting that the surrogate relies on graph substructures similar to those used by the original GNN.

We also visualize the edge attributions produced by both models. We apply Grad to both the original and robust GCNs using the same input graphs. As shown in Figure~\ref{fig:preserve}, the highlighted substructures are highly consistent across both models, with only minor variations in intensity. This consistency confirms that the surrogate model preserves the reasoning logic of the original model, ensuring that the explanations generated via the surrogate are valid for the original target.


\begin{figure}[ht]
    \centering
    \includegraphics[width=1.0\linewidth]{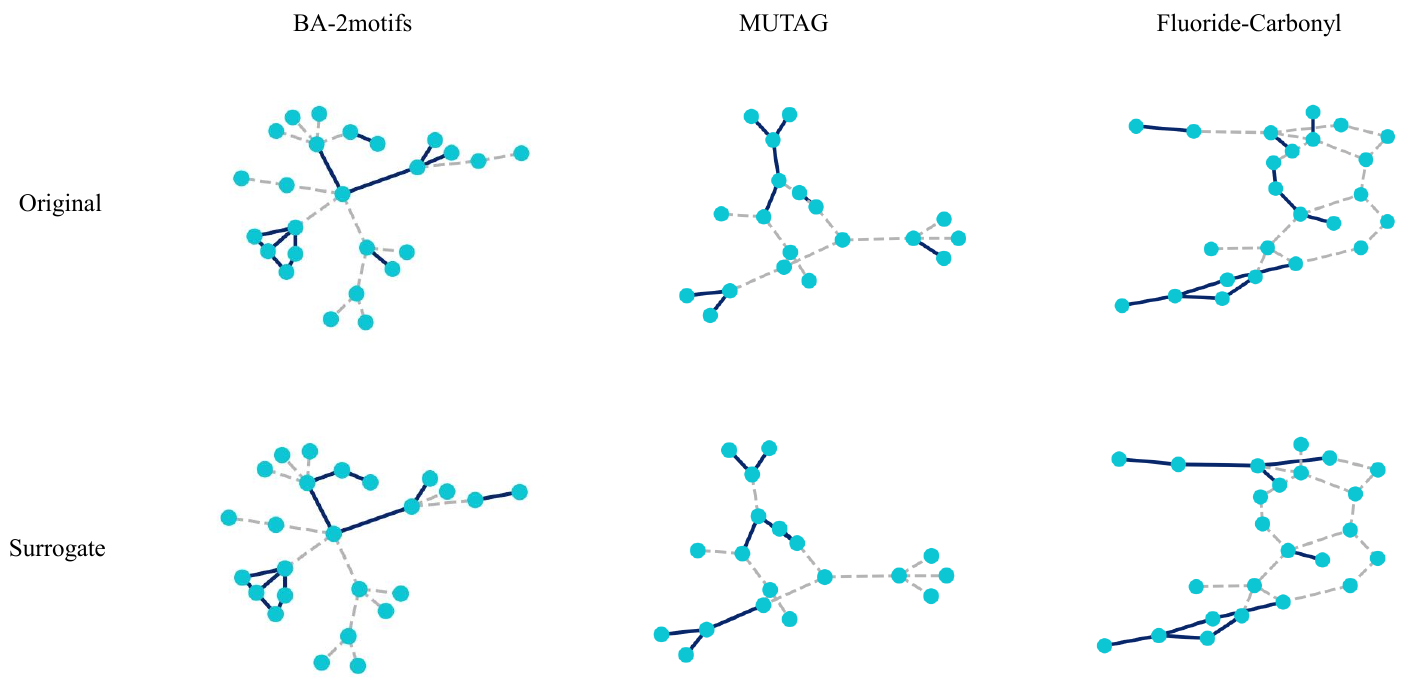}
    \caption{The explanation visualization comparison between original and the robust surrogate GCNs} 
    \label{fig:preserve}
\end{figure}

\subsection{Weight Distribution Visualization}
\label{app:distribution}

To illustrate the edge weight sampling strategy used in our iterative framework, we visualize the distributions of sampled weights for explanatory and non-explanatory edges. Specifically, we draw samples from two truncated Gaussian distributions: \( W_1 \sim \mathcal{TN}(\mu_1, \sigma^2; 0, 1) \) for important edges, and \( W_2 \sim \mathcal{TN}(\mu_2, \sigma^2; 0, 1) \) for less important edges, where \( \mu_1 > \mu_2\). 
Figure~\ref{fig:distirubtion} presents the distributions of sampled edge weights for the {\bamo}, {\mutag}, and {\fluo} datasets. In each subfigure, the blue histogram represents the distribution of weights assigned to non-explanatory edges, and the green histogram corresponds to explanatory edges. We also overlay the histograms with kernel density estimation curves corresponding to the  Gaussian distributions \(\mathcal{N}(\mu_1, \sigma^2) \) and \(\mathcal{N}(\mu_2, \sigma^2) \), shown as solid purple and dashed red curves, respectively. 

\begin{figure}[t]
    \centering
    \includegraphics[width=0.95\linewidth]{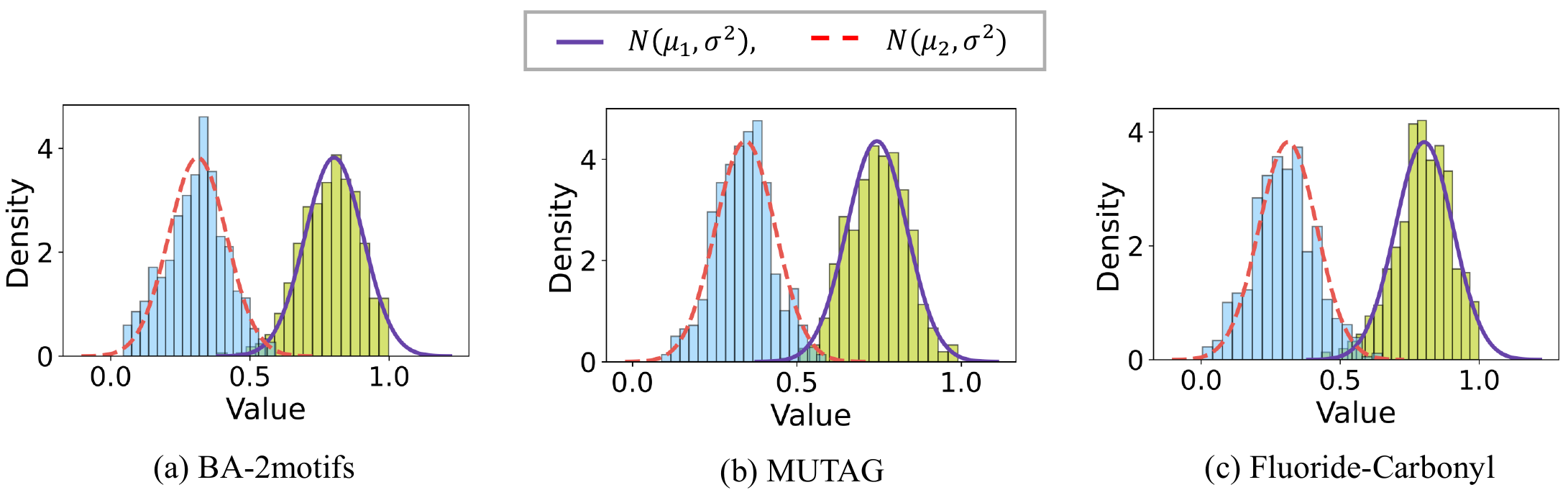}
    \caption{Sampled edge weight distributions for explanatory and non-explanatory edges.}
    \label{fig:distirubtion}
\end{figure}

\begin{figure}[t]
    \centering
    \includegraphics[width=1.0\linewidth]{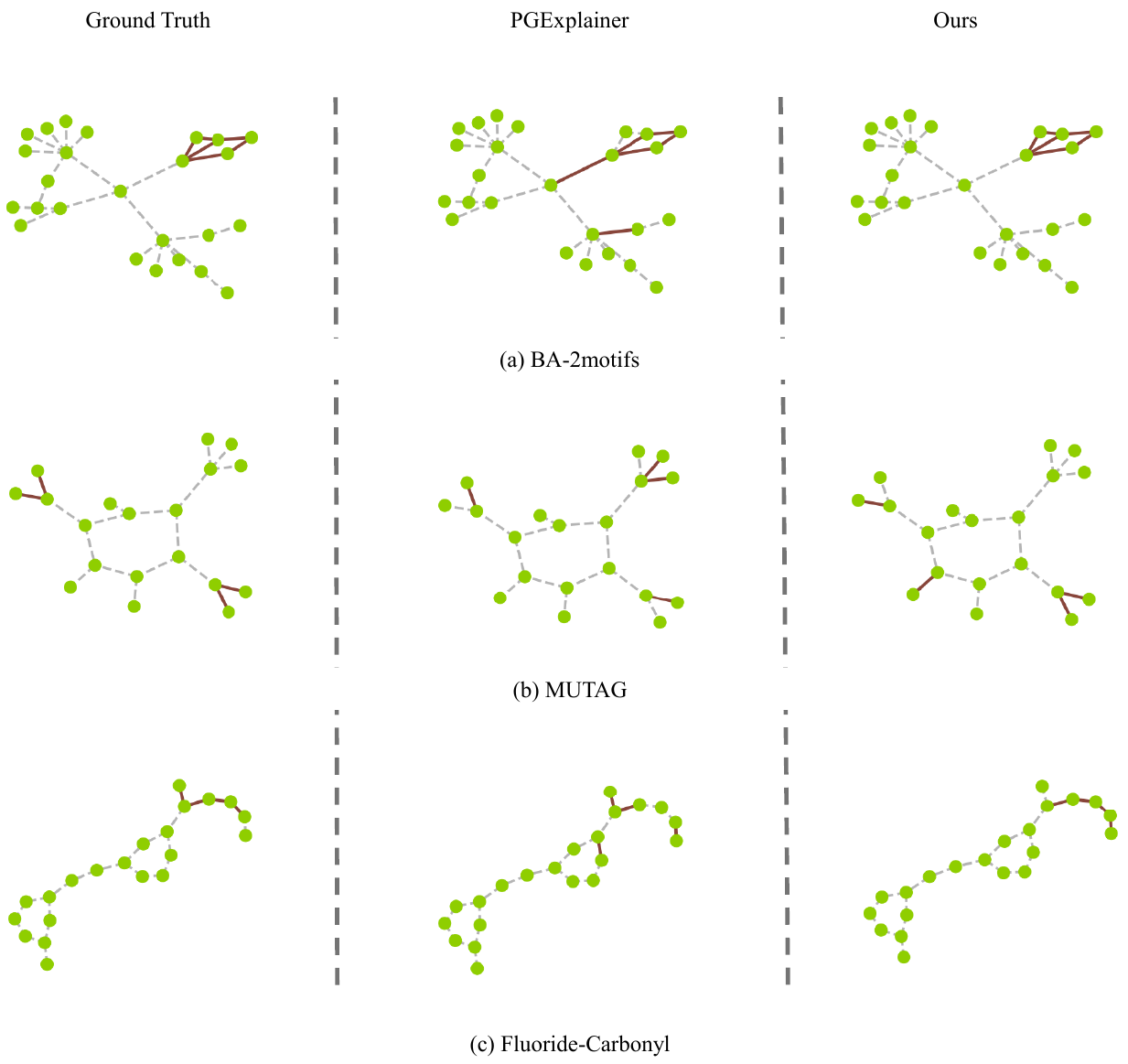}
    \caption{Visual comparison of explanation results.}
    \label{fig:casestudy}
\end{figure}

As shown in the figure, while the actual sampling process is based on truncated Gaussian distributions, the curves demonstrate that the sampled histograms closely resemble the Gaussian distributions. 
In addition, the two distributions exhibit a clear separation. These results confirm that the Gaussian-based sampling strategy enhances structural signals through explanation-aware weighting, and constrains noise from irrelevant edges, enabling the GNN to learn more effectively.

\subsection{Case Study}
\label{app:casestudy}

To qualitatively evaluate the effectiveness of {\ours}, we conduct case studies on representative samples. Figure~\ref{fig:casestudy} visualizes the ground-truth substructures, the explanations produced by the vanilla model, and those generated by our method. In each subfigure, important edges are highlighted in bold brown. 
For fair comparison, we select the same number of top-ranked edges from the explanation as the number of edges in the ground-truth.
From the results, {\ours} produces explanations that more closely match the ground-truth substructures, whereas the vanilla baseline also selects edges outside these regions.

\subsection{Declaration of LLM usage}

We used LLMs only for sentence-level language refinement and grammar polishing. The scientific content, methodology, experiments, and conclusions were developed and verified by the authors.



\end{document}